\pdfoutput=1
\documentclass[10pt,twocolumn,letterpaper]{article}

\usepackage[pagenumbers]{cvpr}

\usepackage{multirow}
\newcommand{\ours}{CoSel}
\newcommand{\ds}{DS}
\newcommand{\rc}{RC}

\definecolor{cvprblue}{rgb}{0.21,0.49,0.74}
\usepackage[pagebackref,breaklinks,colorlinks,allcolors=cvprblue]{hyperref}

\def\confName{CVPR}
\def\confYear{2027}

\title{\ours{}: Cooperation as Selection for End-to-End Driving}

\author{Yiping Li \quad Haohua Que \quad Zhipeng Bao \quad Wenjie Zhao \quad Qianwen Li\\
University of Georgia\\
{\tt\small \{Yiping.Li, Haohua.Que, zb28097, Wenjie.Zhao1, Cami.Li\}@uga.edu}
}

\begin{document}
\maketitle

\begin{abstract}
Cooperative driving can extend a vehicle’s awareness beyond its own sensors, but existing cooperative end-to-end methods inject shared information into the planner and rely on cooperative driving demonstrations to learn how to use it. The moments in which shared information should change a driving decision are rare in such data, providing limited supervision for learning this behavior. We argue that cooperation need not be learned by the planner. Instead, reacting to shared information requires predicting how surrounding agents will move and selecting a response from the candidate trajectories of an ego planner. We introduce \ours{}, a cooperation-as-selection framework that leaves the planner and its ego-only input unchanged. A separately trained cooperative predictor uses observations from the ego vehicle, roadside units, and connected automated vehicles to predict surrounding agents and their future motion. \ours{} evaluates the planner’s candidate trajectories against these predictions, preserves the planner’s choice when it remains safe, and selects the closest safe alternative when shared information reveals additional risk. Each safety intervention is traceable to the shared evidence that triggers it. In closed-loop evaluation, \ours{} achieves a driving score of 70.7 on V2Xverse, outperforming the best compared cooperative method by 7.5 points, and the highest average driving score of 85.8 on MDrive. Under missing ego detections, the collision count of \ours{} increases only from 17 to 19 in 72 scenarios, compared with 19 to 32 for an existing cooperative end-to-end model.
\end{abstract}

\begin{figure}[t]
  \centering
  \includegraphics[width=\linewidth]{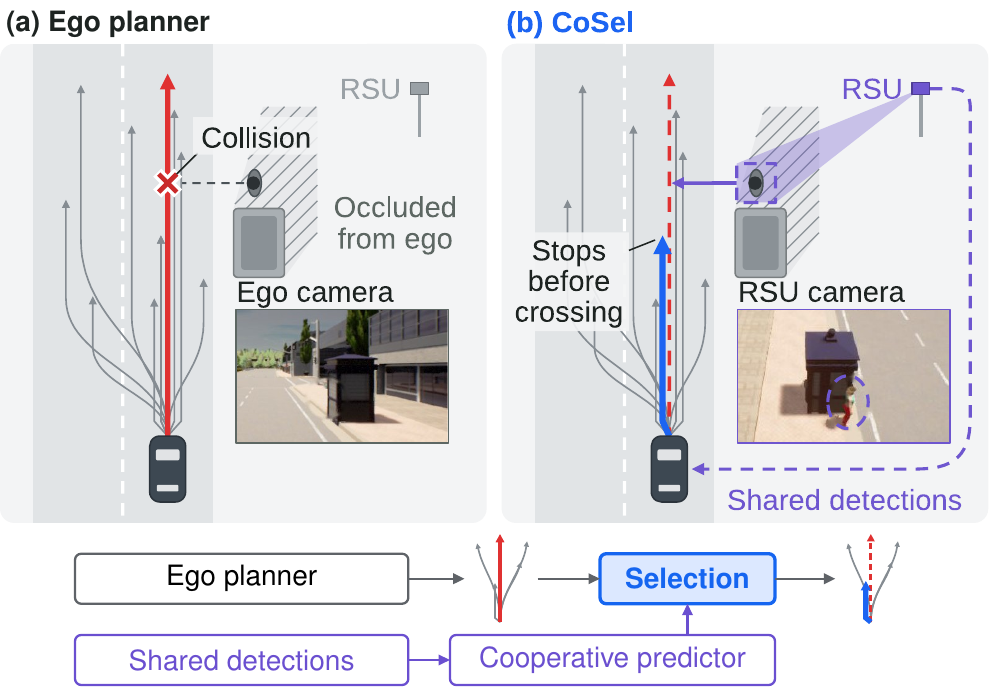}
  \caption{\textbf{Cooperation as selection.} (a) An ego planner cannot see the pedestrian behind the kiosk, and
    its choice (red) leads to a collision. (b) \ours{} keeps the same planner and its candidates (gray). With the detection shared
    by the RSU (purple, dashed), it executes another of these candidates in place of the planner's choice: the
    one on the same path with a slower speed profile, which stops before the crossing (blue).}
  \label{fig:teaser}
\end{figure}

\section{Introduction}
\label{sec:intro}

Cooperative driving extends a vehicle's awareness with the perception~\cite{opv2v,v2xvit,where2comm,dairv2x,v2xseq}
and intent~\cite{colmdriver,codrivingllm} shared by roadside units (RSUs) and other connected automated vehicles
(CAVs).
A single vehicle, however good its end-to-end planner~\cite{sparsedrivev2,lead}, cannot obtain this information from
its own sensors: what lies behind an occluder, such as a pedestrian behind a kiosk (\cref{fig:teaser}), and what other
vehicles intend, as at an unsignalized intersection.

Existing cooperative end-to-end methods fuse the shared information into the planner, as raw data~\cite{v2xvlm} or as
features or queries~\cite{codriving,univ2x,unimmv2x}, and train the planner on cooperative driving data.
Language-based methods instead pass shared messages to a large language
model~\cite{langcoop,codrivingllm,v2xunipool,colmdriver,coopreflect}. In the fusion-based methods the perception that
feeds the planner can be supervised with labels, but how the planner uses it is learned from the expert's trajectory.
The planner has to infer from imitation alone \emph{when} shared information should change its decision.

This is hard to learn, because the frames that could teach it are rare. Imitation supervises a planner with
one ego trajectory per frame, a sparse signal~\cite{chen2024e2esurvey,drivevlaw0}, and the frames in which the expert
changes its action because of an event in the scene, such as a light turning green or a pedestrian stepping out, are
a very small fraction~\cite{codevilla2019exploring,wen2021keyframe}. As a result, a learned planner can reach a low
imitation error by continuing its own speed and past
motion~\cite{codevilla2019exploring,wen2021keyframe,li2024egostatus}, and an additional input is not guaranteed to be
used correctly~\cite{dehaan2019causal}. Of these frames, only a subset can teach the planner anything about shared
information: those in which the event is reported by an RSU or a CAV and is not visible to the ego's own sensors. We
call these frames \emph{decision-changing moments}: only in them should shared information change the decision. In
every other frame it adds little to what the ego already sees, and the demonstration gives no signal for using it.
In V2Xverse training data, 97.7\% of the 7{,}158 brake causes of the expert are visible to the ego, and none is an agent ahead
that only the RSU sees (\cref{sec:suppl_decisive}).
A recent closed-loop benchmark also reports that sharing perception does not
always improve planning~\cite{mdrive}.

We argue that shared information need not enter the planner at all. Two observations support this. First, predicting the future motion of the
surrounding agents from shared detections is \emph{densely supervised}: the future track of every observed agent in every frame is a
training target~\cite{lav}, and an agent that only another source sees is obtained by dropping its ego detection, without a new
expert action. Shared observations have been shown to
improve forecasting~\cite{v2xgraph,cmp,v2xpnp}. Second, an
ego planner that scores candidate trajectories~\cite{vadv2,hydramdp,hydranext,wote,diffusiondrive,goalflow}
typically has the kind of response that a decision-making moment demands. Candidates who follow the same path at different speeds cover slowing and
yielding. Recent vocabularies also vary the path: they are factorized into paths and speed
profiles~\cite{sparsedrivev2}, and candidates ranked below the executed one can include lateral alternatives such as
lane changes~\cite{diffusiondrive}, so taking a neighboring path is also a candidate. The planner lacks only
the information to prefer one of them. Given predicted futures, we can test in every frame whether each candidate is safe,
and when to react follows from this test. We cast \emph{cooperation as selection}: the planner proposes, and shared
information decides which candidate from its vocabulary is executed. Selecting among a learned planner's own candidates with an explicit
safety check has been used for the safety of ego-only planners~\cite{safepathnet,vadv2}. We make it the only place where
cooperation enters.

We propose \ours{}, a cooperative driving system built on this formulation (\cref{fig:method}). Its planner uses
ego-only input, in training and at test time. A cooperative predictor estimates an existence probability and several possible futures for every surrounding agent
from the detections of the ego, RSUs and CAVs. Some inputs enter the same scene with a given future: the plan that a
CAV shares and the static obstacles that a roadside LiDAR sees. Among
the candidates that satisfy a chance constraint on predicted collision risk with a fixed risk budget, a selector
executes the one closest to the planner's choice. Only the predictor is trained on cooperative data,
and its targets are the future tracks of all agents.

Cooperation as selection gives \ours{} three properties by design, where an \emph{intervention} is a step at which
the executed candidate differs from the planner's choice. \emph{Planner first}: the planner's choice is executed
unchanged unless the predicted risk rules it out or progress requires a move (\cref{sec:selection}). \emph{One safety rule for every
source}: an agent or obstacle enters the chance constraint as one factor of the same form whether the ego, an RSU or
a CAV reported it, and a source that reports nothing contributes no factor and leaves the rule unchanged
(\cref{sec:safety,sec:exp:channel}). \emph{Traceable interventions}: every safety intervention is attributed to the
agent or obstacle that contributes most to the predicted risk of the planner's choice, together with the source
that reported it (\cref{sec:exp:qual}).

On the 85 closed-loop routes of V2Xverse~\cite{codriving}, the planner alone reaches a driving score of 57.6,
the selection on the ego's own detections 62.9, and with the detections of the RSU 70.7, the highest among the compared
methods. \ours{} surpasses the cooperative method CoDriving (59.1) and has 67\% fewer collisions per kilometer. \ours{} is robust to unreliable ego perception. In a stress test of 72 scenarios that removes the pedestrians from the ego's sensors,
it collides in 19 scenarios, against 17 without the removal. CoDriving collides in 32, against 19, and the planner alone in 50.
On MDrive~\cite{mdrive}, which tests interaction among multiple CAVs, \ours{} attains the highest
average driving score among the compared methods (85.8, \cref{tab:mdrivemain}). Our contributions are:
\begin{itemize}
  \item \textbf{Formulation.} Cooperation as selection: the planner is trained on ego-only demonstrations, and shared
  information decides which of its candidates is executed.
  \item \textbf{Method.} \ours{}: a lightweight, densely supervised cooperative predictor and an interpretable selector
  that applies one safety rule to every source and leaves the planner untouched.
  \item \textbf{Evidence.} Closed-loop results on V2Xverse and MDrive, a stress test that isolates the shared
  information, and an analysis of each source and of the interventions (\cref{sec:exp:channel,sec:exp:qual}).
\end{itemize}

\section{Related Work}
\label{sec:related}
\paragraph{Cooperative driving.}
Early work on cooperative driving focused on perception: vehicles and roadside units exchange raw data, intermediate features or
detected objects and fuse them to see beyond occlusion, with datasets~\cite{opv2v,dairv2x,v2xseq,v2v4real,v2xreal,tumtrafv2x,rcooper,holovic}
and fusion schemes that are efficient in bandwidth and robust to delay, pose error and
heterogeneity~\cite{v2xvit,where2comm,cobevflow,coalign,codefilling,cosdh,hetecooper,v2xr}. Later work extended cooperation to
prediction~\cite{v2xseq,cmp,comtp,v2xgraph,v2xpnp} and to planning. End-to-end cooperative systems fuse shared features or queries and
then plan, training the whole stack jointly. Most are evaluated in open loop~\cite{univ2x,unie2ev2x,unimmv2x,vips}, while
V2Xverse/CoDriving~\cite{codriving} closes the loop in CARLA. COOPERNAUT~\cite{coopernaut} learns a fused driving policy in
closed loop on scenarios constructed so that the shared view decides the outcome, with corrective demonstrations collected
online. V2Xverse training data rarely contain such moments (\cref{sec:suppl_decisive}). A parallel line shares intents in natural language and negotiates
maneuvers with LLM- or VLM-based agents~\cite{colmdriver,langcoop,codrivingllm,v2vllm,uncap}, and closed-loop benchmarks such as
InterDrive~\cite{colmdriver} and MDrive~\cite{mdrive} now measure multi-vehicle interaction directly. These works differ in what is shared
and where it enters the stack, but in almost all of them the planner is trained, or prompted, on cooperative input and has to learn from
demonstrations when shared information should change its decision. Shared information
is either always used or gated by learned communication or hand-crafted rules~\cite{where2comm,codriving,colmdriver}, and its uncertainty
enters at most as a detection confidence or a calibrated score for an LLM decision~\cite{uncap}. LACO~\cite{laco} is, to our knowledge,
the only training-free plug-in. It exchanges latent states between VLA planners, and its effect on individual decisions is not examined.
We instead cast cooperation as selection: the planner uses ego-only input, shared detections feed a densely supervised
predictor, and one chance constraint over the planner's own candidates decides when shared information changes the decision.

\paragraph{End-to-end and candidate-based planning.}
End-to-end planners map sensors to trajectories, either with a single network that chains perception, prediction and
planning~\cite{uniad,vad,paradrive,sparsedrive}, with generative decoders~\cite{diffusiondrive,goalflow,diffusionplanner}, or with world
models that imagine future observations to supervise or evaluate plans~\cite{drivewm,law,wote}. A growing family of
planners separates trajectory generation from selection: they score a fixed or sampled vocabulary of candidates and execute the best
one~\cite{vadv2,hydramdp,gtrs,ipad,drivesuprim,sparsedrivev2}, mirroring classical planners that rank proposals with hand-crafted or learned
costs~\cite{pdmclosed,dtpp}. Recent work evaluates candidates with world models~\cite{wote} or re-ranks the candidates of a
frozen planner~\cite{mpa}. Rule-based selection among a learned planner's own candidates~\cite{safepathnet,vadv2} and safety filters on
learned policies~\cite{alshiekh2018shield} guard a single vehicle with its own observations. We use selection as the only place
where cooperation enters. Its safety check follows planning under prediction uncertainty, with chance constraints over
multimodal forecasts~\cite{blackmore,racp} and contingency plans that keep a safe fallback~\cite{hardy2013contingency,pek2021failsafe}. We build on SparseDriveV2~\cite{sparsedrivev2}, which scores a factorized vocabulary of paths and speed profiles and exposes both the
candidates and their scores. \ours{} reads both and leaves its network unchanged.

\section{Method}
\label{sec:method}

\begin{figure*}[t]
  \centering
  \includegraphics[width=\linewidth]{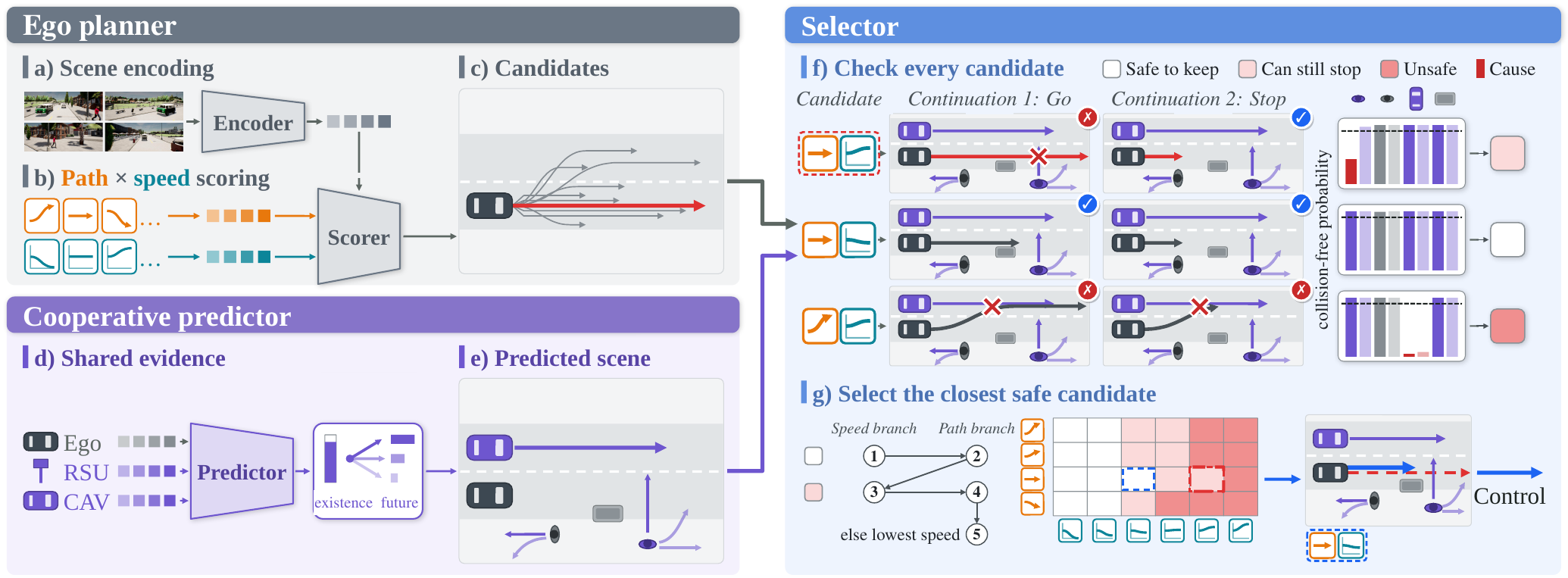}
  \caption{\textbf{Overview of \ours{}.}
    (a--c) The ego planner scores paths and speed profiles from the ego's own cameras and proposes its
    candidates (gray) and its choice (red).
    (d, e) The cooperative predictor takes the detections of the ego, RSUs and CAVs and predicts whether each agent
    exists and how it will move.
    (f) The selector tests every candidate against every agent of the scene for its two continuations: Go, the ego keeps
    executing the candidate, and Stop, the ego brakes to a standstill; a badge marks whether a continuation stays free of a predicted
    collision. A pair of bars belongs to one agent and gives the probability that Go (dark) and Stop (light) stay free of a collision with it.
    Gray bars are agents that the ego detects, purple bars agents that an RSU or a CAV reports, and the dashed line is the required level;
    a red bar marks the agent that caused the intervention.
    (g) The selector executes the candidate (blue dashed) closest to the planner's choice (red dashed) among those that pass for every
    agent, searching in the order 1--5.}
  \label{fig:method}
\end{figure*}

\paragraph{Cooperation as selection.}
\Cref{sec:intro} argued that reacting to shared information takes two things: anticipating how the surrounding agents
will move, and a maneuver that avoids them.
\ours{} takes the maneuvers from an ego
planner (\cref{sec:planner}, \cref{fig:method}a--c) and the futures from a cooperative predictor (\cref{sec:predictor},
\cref{fig:method}d,e), and neither is trained on cooperative demonstrations.
At every decision step, the planner $\pi$ returns a set of candidate trajectories $\mathcal{C}$ and its choice
$c_\pi\in\mathcal{C}$ from the ego's own input.
RSUs send the ego their detections, and other CAVs send their detections, their states and their plans.
Together with the ego's own detections $E^{\text{ego}}$, these form the cooperative evidence $E$.
A selector with no learned parameters (\cref{fig:method}f,g) executes
\begin{equation}
  c^\ast=\operatorname*{arg\,min}_{c\in\mathcal{C}}\, d(c,c_\pi)
  \ \ \text{s.t.}\ \ \Pr(c\ \text{is safe}\mid E)\ge 1-\delta ,
  \label{eq:problem}
\end{equation}
where $d$ measures the departure from the planner's choice, a candidate is safe if the ego can keep executing it over a
horizon $H$ without a predicted collision, and $\delta$ is a budget on the predicted risk (\cref{sec:safety}).
Shared information enters \ours{} only through the constraint of \cref{eq:problem}, by way of the predicted scene. It
never changes the planner's input, vocabulary or scores.
The selector evaluates the constraint for every candidate at every step and intervenes, \ie, executes a candidate other
than $c_\pi$, only when the predicted risk rules out $c_\pi$.
\Cref{eq:problem} alone leaves three cases open: no candidate satisfies the constraint, the planner stays stationary although
moving is safe, and several CAVs meet at a conflict.
\Cref{sec:selection} resolves all three in the same selection step, without changing the planner.

\subsection{Ego Planner}
\label{sec:planner}

Selection needs a planner that exposes its alternatives.
Our planner follows SparseDriveV2~\cite{sparsedrivev2}, which selects its output from a fixed vocabulary of
trajectories, factorized into geometric paths $p\in\mathcal{P}$ and speed profiles $v\in\mathcal{V}$, and scores paths
and profiles from coarse to fine.
A candidate $c=(p,v)$ defines an ego trajectory $\tau_c$, and $c_\pi=(p_\pi,v_\pi)$ is the planner's final choice.

Its vocabulary is dense: on the training data, the mean position error between the expert's trajectory and the closest candidate is
0.06\,m at the median and below 0.23\,m in 99\% of the frames. It contains the responses that a decision-changing moment calls for: slowing and yielding are speed profiles in
$\mathcal{V}$ on the planner's own path, and taking a neighboring path is a path in $\mathcal{P}$.
The planner scores from the ego's input alone and can discard the response to a hazard that it does not see.
We therefore build $\mathcal{C}$ from what the planner keeps before its final round: $p_\pi$ and the $K$
highest-scored paths that the planner keeps, combined with the speed profiles that it keeps after its first scoring
round.
We always add the stop profile of the vocabulary.
Besides $\mathcal{C}$ with its scores and $c_\pi$, the rest of the system reads only the output of the planner's detection
head, $E^{\text{ego}}$, and the solid lane lines from its online map head.

The planner is trained by imitation of the expert driver on V2Xverse~\cite{codriving} training data.
Its input is that of one vehicle: four cameras, the ego speed, its last steering command and the next navigation
point of its route.
Instead of the one vocabulary entry closest to the expert, a scoring round takes as positives all entries that drive like the expert:
speed profiles that travel about as far and as fast as the expert's (within $\eta_s$ and $\eta_v$), and composed candidates whose profile
travels about as far and whose path stays within $\eta_\tau$ of the expert's trajectory $\tau^{\text{e}}$ without a collision; for paths
it keeps the single closest one.
Near standstill, a positive must also agree with the expert's action, to brake or to drive.
With the score $\hat y_x$ of an entry $x$ and the focal loss $\ell$ of SparseDriveV2, a round with positive set $\mathcal{Y}^{+}$
minimizes
\begin{equation}
  \mathcal{L}=\frac{1}{|\mathcal{Y}^{+}|}\sum_{x\in\mathcal{Y}^{+}}\ell(\hat y_x,1)
  +\sum_{x\notin\mathcal{Y}^{+}}\Delta_x\,\ell(\hat y_x,0),
  \label{eq:setpos}
\end{equation}
where the distance $\Delta_x$ of $x$ to the expert weights the negatives, as in SparseDriveV2
(\cref{sec:suppl_method}).

\subsection{Predicted Scene}
\label{sec:predictor}

The selector sees the world only through the predicted scene: a set of elements, each with an existence probability and
one or more futures (\cref{fig:method}d,e).

\paragraph{Evidence.}
The ego holds $E=E^{\text{ego}}\cup E^{\text{I2V}}\cup E^{\text{V2V}}$: its own detections, the detections
of RSUs, and, from every CAV in range, that CAV's detections, state and plan. A detection is a 3D box reported by one source.
Each detection from the last $L$ steps becomes a token with its pose in the ego frame, size, class, score, source and time
lag.
A shared plan, its first 2\,s as ten points transformed into the ego frame, becomes a plan token with its own MLP.
RSU and CAV detections share this format and differ only in the source feature.
The score and the source are inputs for the existence probability $e_i$.

\paragraph{Futures.}
The cooperative predictor $q_\theta$ embeds every token with an MLP and applies a transformer encoder to the set of
tokens~\cite{lee2019settransformer}, without positional encoding because the positions are token features.
An MLP head decodes every token $i$ of the current step into
\begin{equation}
  q_\theta(E)=\big\{e_i,\ (w_{im},\xi_{im})_{m=1}^{M}\big\}_{i},
  \label{eq:pred}
\end{equation}
an existence probability $e_i$ and $M$ future modes with probabilities $w_{im}$ and trajectories $\xi_{im}$ over the
horizon $H$.
At inference, an RSU or CAV token that lies within an association gate of an ego token of the same class is merged
with it, and the agent keeps the prediction of the ego's token.

Some elements come with a given future and enter the scene without the predictor, with $e_i=1$ and a single mode.
A CAV shares its state and its plan: the candidate that it is executing, chosen by its selector at the previous decision step.
A CAV that reports that it is stationary is taken as stationary until its next message, and a moving CAV whose plan ends in a stop
is taken to follow the plan and stay at its end.
Staying and braking need no other agent, so these messages hold whatever the others do; a plan that ends in motion holds only if
the others let the CAV through, so it is only an input token of the predictor (\cref{sec:suppl_method}).
A walking pedestrian is not bound to a lane, so we treat it more defensively: besides its predicted modes, it enters with $e_i=1$
as a disc that keeps its observed velocity over $H$.
A candidate that crosses a static obstacle is not safe.
The detectors miss cones, barriers and debris, which are not among their training classes; on MDrive, each persistent cluster of roadside-LiDAR
points that no detection explains is an obstacle with a stationary mode.
The solid lane lines that the planner's online map predicts ahead of the ego are static obstacles as well, since vehicles must not
cross them, and so, on V2Xverse, are the edges of a corridor toward the current navigation point while the ego starts from a standstill;
every static obstacle enters the scene with $e_i=1$ (\cref{sec:suppl_method}).
The predicted scene is then one set of $N$ elements with existence probabilities and futures, and the source of an element
plays no further role.

\paragraph{Training.}
The predictor is trained on V2Xverse training data with the logged futures of the agents as targets, so every agent in every
frame is a target.
Its inputs are the detections that the detectors output on these data, as at inference, because a predictor trained on
ground-truth boxes loses accuracy on detector outputs~\cite{xu2024realworld}.
A set of detections can be edited without invalidating its target: we remove the ego's detections of an agent that another source also sees
(\emph{counterfactual occlusion}), and we drop whole sources at random.
The predictor therefore meets in training an agent that only an RSU or a CAV reports, and a source that sends nothing.
\Cref{sec:exp:channel} tests the first case with the pedestrians removed from the ego's sensors and \cref{sec:exp:abl} the second with a source switched off.

\subsection{Safety Probability of a Candidate}
\label{sec:safety}

We compute the probability in \cref{eq:problem} element by element.

\paragraph{Probability.}
Let $s_c(i,m)\in\{0,1\}$ indicate that the ego has no predicted collision with mode $m$ of element $i$ within $H$ when it keeps
executing candidate $c$, \ie, that their footprints do not overlap.
A candidate lasts 2\,s, the planner's horizon; to catch later conflicts, the check runs over $H$ with the ego going straight at its
last speed after the candidate ends.
Every predicted collision counts, including when the ego is stationary: a candidate that stays in the predicted path of an agent is not safe
to keep.
Treating the elements as independent,
\begin{equation}
  P_c^{\text{go}}=\prod_{i=1}^{N}\Big[(1-e_i)+e_i\sum_{m=1}^{M}w_{im}\,s_c(i,m)\Big]
  \label{eq:pb}
\end{equation}
is the probability that $c$ can be kept, the probability in \cref{eq:problem}.
Independence is an approximation, since interacting agents are correlated.
On the validation Towns 7 and 10, unseen maps for the predictor like Town05, the resulting risk errs on the safe side: for constant-speed
probe trajectories a predicted risk $1-P_c^{\text{go}}$ of 0.35 corresponds to an observed collision frequency of 0.24, and the budget
$\delta$ of \cref{sec:impl} to an observed risk of about 0.2 (\cref{sec:suppl_calib}).

\paragraph{One safety rule for every source.}
The source of a detection is an input of the predictor and does not appear in \cref{eq:pb}.
An agent or obstacle is one factor of the same form whether the ego, an RSU or a CAV reported it: an agent detected
only by an RSU adds one factor, and an element with $e_i=1$ and a single mode has a factor of $0$ or $1$.
A source that sends nothing contributes no factor, and without messages the selector applies the same constraint to
the ego's own detections.
Sources can be added or removed without changing the rule (\cref{sec:exp:channel}).

\paragraph{Risk budget.}
$\delta$ is a threshold on the predictor's own probabilities, applied at every decision step.
It is needed because the predictor's error grows with the horizon: on the validation towns, its best-mode error averages
$0.5$\,m over the first 2\,s and reaches $2.0$\,m at 4\,s, more than half a lane width (\cref{sec:suppl_calib}).
Near an intersection, the modes at the end of the horizon cover much of the road, and a requirement to avoid every
mode rejects most moving candidates.
The chance constraint accepts a candidate that conflicts only with modes of low probability, and the next decision
step tests it again.
We use one $\delta$ in all experiments (\cref{sec:impl}); a smaller $\delta$ is more cautious (\cref{sec:suppl_delta}).

\subsection{Cooperative Selector}
\label{sec:selection}

\paragraph{Selection.}
Let $\mathcal{G}=\{c:P_c^{\text{go}}\ge 1-\delta\}$, with $P_c^{\text{go}}$ of \cref{eq:pb}, be the candidates that are safe to keep.
When $\mathcal{G}$ is empty, \cref{eq:problem} has no solution, and we relax it as in contingency
planning~\cite{hardy2013contingency}.
A new decision takes effect only after a committed prefix $T_p$, one decision period plus the time a new decision
needs to reach the actuators, so a candidate remains acceptable if the ego can still brake to a standstill safely
after this prefix.
Let $P_c^{\text{adm}}$ be the probability that at least one continuation of $c$ is safe: its \emph{go continuation}, which keeps executing $c$, or its \emph{stop continuation},
which brakes to a standstill after $T_p$ (\cref{sec:suppl_method}), and let the \emph{admissible} candidates be
$\mathcal{A}=\{c:P_c^{\text{adm}}\ge 1-\delta\}\supseteq\mathcal{G}$.
The ego chooses between the two at the next decision steps, so $P_c^{\text{adm}}$ is an upper bound for a committed choice
(\cref{sec:exp:abl} ablates the stricter rule).
The selector executes
\begin{align}
  k(c)&=\big(\mathbf{1}[c\notin\mathcal{A}],\ \mathbf{1}[c\notin\mathcal{G}],\ d(c,c_\pi)\big), \nonumber\\
  c^\ast&=
  \begin{cases}
    \operatorname*{lex\,min}\limits_{c\in\mathcal{C}}\ k(c) & \text{if }\mathcal{A}\neq\emptyset,\\[3pt]
    \operatorname*{arg\,min}\limits_{c\in\mathcal{C}}\ v_c(T_p) & \text{otherwise.}
  \end{cases}
  \label{eq:select}
\end{align}
This follows minimum-violation planning~\cite{tumova2013least,censi2019rulebooks}: the constraints are ordered, and one that
cannot be met gives way while the others still hold.
In the second case no candidate keeps a safe continuation, and the selector executes the candidate with the lowest
ego speed $v_c(T_p)$ at the end of the committed prefix, with ties broken by $d$, a minimal-risk
maneuver~\cite{pek2021failsafe}. This maneuver lowers the speed and does not minimize the predicted risk.
After this safety rule, $d$ splits the selection into the two branches of the vocabulary: the planner's own path first, then the
nearest path, then the nearest speed profile.
\emph{Speed branch}: when $c_\pi$ fails the rule, the selector keeps $p_\pi$ and takes the nearest profile that passes, usually a slower
one; when the planner has stayed stationary for $T_{\text{stall}}$ while a moving profile on $p_\pi$ stayed in $\mathcal{G}$, it takes the
moving profile that the planner scores highest.
\emph{Path branch}: the selector takes another path only when no profile on $p_\pi$ passes, or when a static obstacle blocks $p_\pi$
within a look-ahead distance $D$; the vocabulary paths that pass the obstacle then join $\mathcal{C}$ and rank above the blocked path if
one of them is in $\mathcal{G}$ (\cref{sec:suppl_method}; \cref{sec:suppl_alg} lists one decision step in full).
The stall and the blocked path are progress cases: they rank below safety, as in rulebooks~\cite{censi2019rulebooks}, and choose only
among candidates in $\mathcal{G}$, so they shorten waits without lowering safety.
On Bench2Drive the selector improves the released SparseDriveV2 with both progress cases disabled (\cref{sec:suppl_b2d}).

\paragraph{Properties.}
\Cref{eq:pb,eq:select} give \ours{} the first and third properties of \cref{sec:intro}, and \cref{sec:safety} showed
the second.
\emph{Planner first}: if $c_\pi\in\mathcal{G}$, the planner's choice is executed unchanged, unless one of the
two progress cases applies.
\emph{Traceable interventions}: a change made by the first two terms of $k$ is a \emph{safety} intervention, and
because \cref{eq:pb} is a product, we report the factor of $P_{c_\pi}^{\text{go}}$ with the smallest value, the dominant
contributor to the risk, and the source that reported it.
When several factors are needed to bring $P_{c_\pi}^{\text{go}}$ below $1-\delta$, this names the largest contributor and is not
a causal attribution.

\paragraph{Several CAVs: yielding as one more factor.}
At an unsignalized intersection, the selections of several CAVs are coupled, and two cautious CAVs can wait for each
other indefinitely.
We decompose the joint problem with one shared priority order, as in prioritized planning~\cite{cap2015prioritized},
and let the order act through the scene.
Two CAVs are in conflict when their paths ahead come closer than a radius $r_c$ and one does not follow the other; the \emph{conflict zone}
of the CAV with priority is a 6.9\,m by 4\,m box on its path where the two paths meet.
A lower-priority CAV adds this zone, as an obstacle,
to its scene with $e_i=1$ and a stationary mode until that CAV has passed, as in reservation-based intersection
management~\cite{dresner2008aim}.
Yielding is then one more factor of \cref{eq:pb}, and I2V and V2V cooperation share one mechanism.
To derive the order, CAVs also share their upcoming navigation points, which give each CAV's path ahead up to $D$ and
show a turn before the CAV starts to make it.
All CAVs derive the same total order from the shared states, with the earlier estimated arrival at the conflict first
and ties broken by the CAV index, and the order of a pair is fixed while their conflict lasts, unless the CAV with
priority cannot move within a time-out $T_{\text{exp}}$ or has arrived (\cref{sec:suppl_method}).
A yielding CAV that is stationary and has no candidate in $\mathcal{G}$ stays stationary.
We assume that CAV states arrive at every decision step and do not study message delay
or loss.

\paragraph{What is learned.}
The planner imitates ego-only demonstrations, and the predictor regresses the logged futures of all agents. The selector has no learned
parameters, and the system is not trained end to end.
Decision-changing moments enter \ours{} only through the constraint of \cref{eq:problem}, which the selector evaluates at every step.

\section{Experiments}
\label{sec:exp}

\begin{figure*}[t]
\centering
\includegraphics[width=\textwidth]{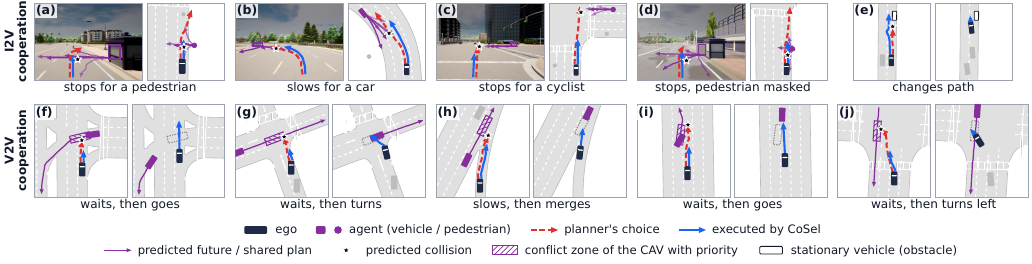}
\caption{\textbf{What the selection changes.} I2V cooperation: an RSU reports an agent that the ego does not detect, and the ego stops or slows
(a--d; in (d) the pedestrian is masked from the ego's sensors). In (e) the roadside LiDAR reports a stationary vehicle in the planner's path, and the ego
takes another path. V2V cooperation: the ego waits or slows before the conflict zone of a CAV with priority and goes once that CAV has passed.
In (e)--(j), left: the decision; right: afterward. The star marks where the planner's choice is predicted to collide.}
\label{fig:instances}
\end{figure*}

We evaluate \ours{} in closed loop on V2Xverse (one vehicle, one RSU) and MDrive (several CAVs) (\cref{sec:exp:main}),
stress-test it on 72 short scenarios that we stage (\cref{sec:exp:channel}) and ablate its parts (\cref{sec:exp:abl}).
\Cref{fig:instances} shows qualitative results: decisions that shared information changes.

\subsection{Setup}
\label{sec:exp:setup}

\paragraph{Benchmarks and metrics.}
\emph{V2Xverse-85}~\cite{codriving} is the closed-loop benchmark of the V2Xverse platform: 85 route--scenario pairs on 58
routes in CARLA Town05, with an RSU placed 12\,m ahead of the ego on the right side of the road, 7.5\,m high.
\emph{MDrive}~\cite{mdrive} controls up to eight CAVs in one scene, and we use its three subsets InterDrive (46 scenarios),
Interaction (112) and V2XPnP (67).
We report the driving score (\ds{}), route completion (\rc{}), infraction score (IS) and collisions per kilometer on V2Xverse, as
in~\cite{codriving}, and \ds{} and success rate (SR) on MDrive, as in~\cite{mdrive}.
A difference between two methods is paired over the same routes or scenarios and is given with a bootstrap 95\%
confidence interval over pairs or scenarios. Every method is run once.

\paragraph{Baselines.}
On V2Xverse-85 we run every baseline of \cref{tab:main} ourselves, with its released weights and on the same routes. On MDrive we list the
numbers published in~\cite{mdrive}.
\emph{\ours{} planner only} is our planner alone.

\paragraph{Implementation.}
\label{sec:impl}
The planner is trained on ego-only expert data of V2Xverse, recorded with two vehicles, and the predictor on its data with two vehicles and RSUs, both from Towns
1--4 and 6. Towns 7 and 10 serve for validation, and Town05, where V2Xverse is evaluated, for neither.
The predictor outputs $M=6$ modes over $H=4$\,s.
The risk budget $\delta=0.3$ of the selector was chosen on validation scenarios in Towns 7 and 10
(\cref{sec:suppl_delta}). The committed prefix $T_p=0.4$\,s is one decision period plus one actuation step,
$T_{\text{stall}}$ and $T_{\text{exp}}$ are as long as the horizon, and the start distance $D_s$ is the spacing of the navigation points (\cref{sec:suppl_method}).
The other constants are geometric tolerances, decelerations and thresholds set by hand, and on the validation scenarios the selection is insensitive to
them: varying nine of them changes the share of unsafe steps by less than 0.2 points (\cref{sec:suppl_delta}).
The planner runs on the GPU, and the selector with its two predictor calls can run on a CPU: in an offline replay of six routes
of V2Xverse-85 on a CPU, one selection takes a median of 27--48\,ms, within the decision period of 200\,ms.

\subsection{Comparison with the state of the art}
\label{sec:exp:main}

\begin{table}[t]
\centering\footnotesize
\caption{\textbf{V2Xverse-85}, every method run by us on the same 85 route--scenario pairs. Ped / Veh / Lay: collisions
per km. Bold: best in each group.}
\label{tab:main}
\setlength{\tabcolsep}{3.4pt}
\begin{tabular}{l ccc ccc}
\toprule
Method & \ds$\uparrow$ & \rc$\uparrow$ & IS$\uparrow$ & Ped$\downarrow$ & Veh$\downarrow$ & Lay$\downarrow$ \\
\midrule
\multicolumn{7}{l}{\emph{Ego-only}} \\
CoDriving No fusion ~\cite{codriving} & 55.8 & \textbf{93.8} & 0.58 & 3.99 & 11.54 & 6.21 \\
TCP~\cite{tcp} & \textbf{66.7} & 80.7 & \textbf{0.83} & \textbf{2.71} & 2.88 & \textbf{0.51} \\
\ours{} planner only & 57.6 & 79.8 & 0.75 & 5.12 & \textbf{2.39} & 1.88 \\
\midrule
\multicolumn{7}{l}{\emph{Cooperative}} \\
CoDriving Late fusion~\cite{codriving} & 63.2 & \textbf{96.3} & 0.65 & \textbf{2.85} & 8.15 & 5.43 \\
CoDriving Early fusion~\cite{codriving} & 56.7 & 92.9 & 0.60 & 3.88 & 12.25 & 2.54 \\
F-Cooper~\cite{fcooper} & 50.8 & 93.5 & 0.54 & 6.40 & 9.59 & 6.95 \\
V2X-ViT~\cite{v2xvit} & 52.0 & 85.9 & 0.59 & 7.91 & 6.57 & \textbf{0.60} \\
CoDriving~\cite{codriving} & 59.1 & \textbf{96.3} & 0.61 & 4.21 & 9.11 & 5.98 \\
\ours{} & \textbf{70.7} & 88.7 & \textbf{0.79} & 3.04 & \textbf{1.45} & 1.88 \\
\bottomrule
\end{tabular}
\end{table}

\begin{table}[t]
\centering\footnotesize
\caption{\textbf{MDrive}~\cite{mdrive}: \ds{} / SR per subset and their mean. Baselines as published
in~\cite{mdrive}, rounded to one decimal; \ours{} run by us. $^\ast$rule-based variant. Bold: best in each group.}
\label{tab:mdrivemain}
\setlength{\tabcolsep}{2.2pt}
\begin{tabular}{l cc cc cc cc}
\toprule
 & \multicolumn{2}{c}{InterDrive} & \multicolumn{2}{c}{Interaction} & \multicolumn{2}{c}{V2XPnP} & \multicolumn{2}{c}{Mean} \\
\cmidrule(lr){2-3}\cmidrule(lr){4-5}\cmidrule(lr){6-7}\cmidrule(l){8-9}
Method & \ds & SR & \ds & SR & \ds & SR & \ds & SR \\
\midrule
\multicolumn{9}{l}{\emph{Ego-only}} \\
TCP~\cite{tcp} & 83.6 & 67.8 & 75.8 & 55.3 & 71.0 & 58.7 & 76.8 & 60.6 \\
UniAD~\cite{uniad} & 57.5 & 26.4 & 53.1 & 31.1 & 72.6 & 68.5 & 61.0 & 42.0 \\
VAD~\cite{vad} & 57.5 & 23.3 & 48.8 & 27.1 & 67.2 & 66.3 & 57.8 & 38.9 \\
LMDrive~\cite{lmdrive} & 56.7 & 27.1 & 57.5 & 29.1 & 71.6 & 69.6 & 61.9 & 41.9 \\
\ours{} planner only & 86.0 & 78.8 & 76.9 & 65.6 & 83.7 & 71 & 82.2 & 72 \\
\midrule
\multicolumn{9}{l}{\emph{Cooperative}} \\
CoDriving~\cite{codriving} & 84.0 & 66.1 & 79.1 & \textbf{70.5} & 85.0 & 82.8 & 82.7 & 73.2 \\
CoLMDriver$^\ast$~\cite{colmdriver} & 83.8 & 75.1 & 66.7 & 53.7 & 77.9 & 73.1 & 76.1 & 67.3 \\
CoLMDriver~\cite{colmdriver} & 85.3 & 76.9 & 67.9 & 51.9 & 78.3 & 75.4 & 77.2 & 68.1 \\
\ours{} & \textbf{89.6} & \textbf{81.1} & \textbf{80.5} & 68.2 & \textbf{87.3} & \textbf{84.9} & \textbf{85.8} & \textbf{78.1} \\
\bottomrule
\end{tabular}
\end{table}

\Cref{tab:main} compares the methods on V2Xverse-85.
\ours{} has the highest \ds{} and the fewest vehicle collisions per kilometer of all methods, and the highest infraction score among the cooperative methods. The paired \ds{} difference to
CoDriving is $+11.6$ $[+3.7,+19.6]$.
TCP has the highest infraction score and the fewest pedestrian and layout collisions per kilometer, but it ends blocked or timed out on 41 of the
85 routes, \ours{} on 18 (exact McNemar test, $p<0.001$).
\label{sec:exp:qual}%
In 19 of the 31 collisions of \ours{} with agents the ego is hit from behind or while it
is stationary, and 13 of its 17 blocked routes end behind stationary traffic (\cref{sec:suppl_fault}).
In 303 episodes of at least 0.6\,s, agents that only the RSU reports put the planner's choice over the risk budget.
In 73\% of them the ego's own detections confirm the conflict within 6\,s, a median of 1.4\,s later, so the RSU warns earlier.

\Cref{tab:mdrivemain} tests whether our method carries over to several CAVs: on MDrive \ours{} runs with the V2Xverse weights, and the
plans of the other CAVs and their order enter the same scene.
\ours{} has the highest \ds{} on all three subsets and the highest mean SR.
Without the messages of the other CAVs (detections, plans and the order), 21 of the 175 CAVs of InterDrive have a vehicle collision,
and with them 8: 7.4 fewer per 100 CAVs ($[1.1,13.9]$ over scenarios).

\subsection{Stress test}
\label{sec:exp:channel}

\paragraph{Safety-critical scenarios.}
On these routes a hazard usually appears far ahead: when a pedestrian starts to walk toward the ego's path, the ego is a median of 35\,m
from the crossing point on V2Xverse-85 and 20\,m on MDrive.
We therefore stage 72 scenarios in Town05 next to a fixed RSU (\cref{sec:suppl_channel}), where this distance is 12\,m.
In 40 scenarios a pedestrian steps out from behind an occluder 1.4--5.0\,s before it reaches the ego's path, which tests the reaction to a sudden
hazard. In eight control scenarios a pedestrian walks away along the sidewalk, which tests that an irrelevant pedestrian does not slow the ego.
\ours{} collides in 17 of the 72 scenarios and CoDriving in 19.
Most of these collisions leave no time to stop: 9 of the 11 step-out collisions of \ours{} fall in the 18 scenarios in which the pedestrian
enters the ego's lane about 1\,s after it becomes visible, less than a full stop from the approach speed takes (\cref{sec:suppl_channel}).
On the control scenarios \ours{} is as fast as its planner alone and faster than CoDriving (13.9, 13.8 and 15.1\,s): the disc of a
pedestrian who walks away stays off the ego's path.

\paragraph{Robustness to unreliable ego perception.}
A mask removes every pedestrian from the ego's camera images and
LiDAR points for the whole scenario, so that the RSU is the only source that reports them.
The mask is the same for every method, and no method saw masked sensor data in training.
With and without the mask, \ours{} collides in 19 and 17 scenarios (\cref{fig:mask_full}) and CoDriving, which receives the RSU features, in
32 and 19; the planner alone collides in 50 under the mask.
\ours{} is robust to unreliable ego perception because it handles a pedestrian in the same way whichever source reports it
(\cref{sec:safety}).

\subsection{Ablation study}
\label{sec:exp:abl}

\begin{table}[t]
\centering\footnotesize
\caption{\textbf{Ablation on V2Xverse-85.} Each pair of columns gives the result without and with one part of \ours{}, with the same
weights. Blocked: routes blocked or timed out, of 85. $^\ast$Without: constant-velocity futures. $^\dagger$Without: a candidate is kept
only if its go continuation is safe.}
\label{tab:ablation}
\setlength{\tabcolsep}{3.4pt}
\begin{tabular}{l cc cc cc}
\toprule
 & \multicolumn{2}{c}{RSU detections} & \multicolumn{2}{c}{Predictor$^\ast$} & \multicolumn{2}{c}{Stop continuation$^\dagger$} \\
\cmidrule(lr){2-3}\cmidrule(lr){4-5}\cmidrule(l){6-7}
 & w/o & with & w/o & with & w/o & with \\
\midrule
\ds{} $\uparrow$ & 62.9 & 70.7 & 66.5 & 70.7 & 66.9 & 70.7 \\
Blocked $\downarrow$ & 28 & 18 & 21 & 18 & 30 & 18 \\
\bottomrule
\end{tabular}
\end{table}

\paragraph{Shared information.}
\label{sec:exp:ablation}
Without the RSU detections (\cref{tab:ablation}), the selector tests the planner's candidates against the ego's own detections only; this already
raises the \ds{} of the planner from 57.6 to 62.9.
The RSU detections raise it to 70.7 (paired difference $+7.8$ $[+0.3,+15.5]$), with 18 routes blocked or timed out instead of 28 (exact
McNemar test, $p=0.031$) and, in the masked stress test, 19 scenarios with a collision instead of 51.
The roadside LiDAR of MDrive, which its cooperative baselines receive as well, reports the static obstacles that detectors miss: 5.7 of 100
CAVs on Interaction hit such an object with the planner alone and 0.9 with \ours{}.

\paragraph{Safety rule.}
With constant-velocity futures in place of the predictor, 8.9\% of the validation trajectories with zero predicted risk hit an
agent, against 2.8\%, and the \ds{} on V2Xverse-85 is 66.5 instead of 70.7.
Without the rule for walking pedestrians (\cref{sec:predictor}), 26 instead of 19 scenarios of the masked stress test end in a collision, and the \ds{} on them is 80.2
instead of 85.3 (paired difference $+5.1$ $[+0.2,+10.3]$).

\paragraph{Progress.}
Without the stop continuation, 30 routes instead of 18 end blocked or timed out (exact McNemar test, $p=0.017$).

\paragraph{Other planners.}
We also test whether the selector generalizes to other planners or benchmarks. On the released CoDriving model, choosing among 64 velocity profiles, it reaches a \ds{} of 65.6 on V2Xverse-85, against 59.1 for the original (\cref{sec:suppl_host}). On SparseDriveV2~\cite{sparsedrivev2} in Bench2Drive, it raises the \ds{} by $+9.2$ $[+2.0,+17.0]$ over 45 crossing and junction routes (\cref{sec:suppl_b2d}).

\section{Conclusion}
\label{sec:conclusion}

We presented \ours{}, which casts cooperative driving as selection rather than planner retraining. An ego planner proposes candidate trajectories from ego-only input, while information shared by RSUs and CAVs is used to predict the surrounding scene and determine which candidate should be executed. \ours{} preserves the planner's choice when it remains safe and selects the closest safe alternative when shared information reveals additional risk. This separation allows cooperation to benefit from densely supervised prediction without requiring the planner to learn from rare cooperative decision-changing demonstrations.

Without retraining the underlying planner, \ours{} achieves the highest driving scores among the compared methods on both V2Xverse and MDrive. It also remains robust when ego detections are removed, with only a small increase in collisions as shared information becomes the primary evidence of otherwise unseen hazards. These results demonstrate that shared information does not need to be fused into a learned driving policy to improve cooperative driving. Instead, cooperation can complement an existing planner by deciding when to keep its choice and when to select another of its maneuvers. \ours{} has three limitations that define our next steps. First, it has not yet been tested with other state-of-the-art planners, such as DrivoR~\cite{drivor}, which run on different platforms. Second, its risk treats agents as independent and its predictor reads only detections, so we will study predictors of the joint future of interacting agents that also include traffic lights. Third, it assumes ideal communication; message delay, pose error, and real-world V2X data remain to be tested.

\section{LLM Usage Statement}
\label{sec:llm_statement}

We used Claude (Anthropic) throughout this project as an assistive tool. Specifically, it helped with early-stage literature comparison, writing and debugging training/evaluation/cluster scripts, and debugging model modules (the modified single planner, cooperative predictor, and selector). During writing, it was used for grammar checking, language polishing, and statistical significance analysis. We carefully reviewed and verified all AI-generated code and text, and take full responsibility for the content of the paper.

{
    \small
    \bibliographystyle{ieeenat_fullname}
    \bibliography{main,ref}
}

\clearpage
\maketitlesupplementary

\section{Cooperation-decisive events in driving data}
\label{sec:suppl_decisive}
\label{app:census}

An imitation-trained planner can learn to act on cooperative evidence only if its training data contain frames in which that evidence
explains the expert's action. We count such frames in V2Xverse training data. We collected these data with the routes and the spawn positions of the data-collection scripts that
V2Xverse provides, and use them for training and for this analysis.

\paragraph{Data.}
The data cover 167 routes in CARLA~0.9.10, each driven by two ego vehicles under the V2Xverse expert, with road-side units placed as in
V2Xverse (one per ego, re-placed ahead of it as it drives), saved at 5\,Hz. For 18 routes the ego's front segmentation is missing, so its
visibility cannot be determined; we exclude them and keep 149 routes (292 ego runs).

\paragraph{Brake causes.}
The expert brakes when one of its hazard tests fires, and the data record which agents triggered it (pedestrian, bicycle, vehicle ahead,
vehicle in the lane, vehicle in the junction). An \emph{event} is a maximal run of frames in which the same agent is a brake cause for the
same ego; gaps of up to two frames are merged.

\paragraph{Visibility.}
An agent is visible to the ego if at least four corners of its 3D box are visible in the semantic segmentation of the ego's front, left and
right cameras (behind the ego, we use the ego's LiDAR returns as a proxy). It is visible to an RSU if at least four corners are unoccluded in
the depth images of the RSU's four cameras. An event is \emph{RSU-decisive} if, at its first frame, the cause is not visible to the ego but
visible to an RSU. It is \emph{sustained} if the expert brakes for the agent for at least 1\,s and the agent is visible only to the RSU in at
least three of those frames. Events whose cause is the other ego vehicle are not counted as RSU-decisive.

\paragraph{Results.}
Table~\ref{tab:census} summarizes the counts. Almost every brake cause is visible to the ego when braking starts (97.7\%). Of the 106
RSU-decisive events, the 36 in front of the ego are all single-frame visibility flicker: the ego had seen the same agent at least 4\,s
earlier. Only 33 events in 26 scenes are sustained, after removing two whose cause is at an implausible height (\eg a pedestrian below
the road surface that the expert's 2D hazard test still reacts to). All of them are behind or beside the ego: vehicles (22) and bicycles (9)
approaching from behind, mostly while the ego is stopped, and two pedestrians beside and behind it. In both pedestrian events, another
pedestrian ahead and visible to the ego is a brake cause in every frame, so the RSU does not change the expert's action.
The training data therefore contain \textbf{no} event in which the expert brakes for an agent ahead that only the RSU sees.

\begin{table}[t]
  \centering
  \small
  \begin{tabular}{lr}
    \toprule
    Brake-cause events (149 routes, 292 ego runs) & 7{,}158 \\
    \quad vehicle / bicycle / pedestrian & 4{,}653 / 2{,}048 / 457 \\
    \quad cause visible to the ego at onset & 6{,}992 (97.7\%) \\
    \quad cause visible to neither ego nor RSU at onset & 60 \\
    \quad RSU-decisive at onset & 106 \\
    \qquad ahead / behind / beside the ego & 36 / 47 / 23 \\
    \qquad ahead and not single-frame flicker & 0 \\
    Sustained RSU-decisive events (26 scenes) & 33 \\
    \quad vehicles / bicycles behind the ego & 22 / 9 \\
    \quad pedestrians beside or behind the ego & 2 \\
    \quad ahead of the ego & \textbf{0} \\
    \bottomrule
  \end{tabular}
  \caption{Brake-cause events in V2Xverse training data, by who can see the cause when the expert starts braking.}
  \label{tab:census}
\end{table}

\paragraph{Robustness to the visibility labels.}
The RSU visibility labels are approximate: the depth test treats a corner as unoccluded unless the depth image is more than 1\,m closer, so
corners just behind an occluder count as visible, which can only add RSU-decisive events. To make sure the conclusion does not depend on
these labels, we drop them and ask directly whether any forward brake has no cause the ego can see. Of the 1{,}092 sustained ($\geq$1\,s)
forward brake-cause events, none contains three or more frames in which the cause is invisible to the ego and no other ego-visible cause
exists.

\paragraph{Scope.}
Visibility is not detection: an agent visible to the ego can still be missed by its detector, so the count bounds what the data could teach
rather than what a trained planner perceives. The absence of forward RSU-decisive events does not prove that retraining cannot work; it shows
that this data offers no supervision for it, which is why retraining does not guarantee the planner learns to act on such evidence. The
census covers the data we collected; we did not repeat it on the released V2Xverse training set.

\section{MDrive infractions}
\label{sec:suppl_mdrive}
\Cref{tab:mdrive} splits the MDrive results of \ours{} and of its planner alone by infraction.
On InterDrive and Interaction a CAV waits before the zone reserved for another CAV in 409 events across 110 of the 158 scenarios; without the messages of the other CAVs (their detections, plans and the order), 21 of the 175 CAVs of InterDrive have a vehicle collision, against 8
with them, 7.4 fewer per 100 CAVs ($[1.1,13.9]$, bootstrap over the 46 scenarios).
On both subsets the collisions with static objects fall (5.7 to 0.9 and 3.8 to 0.0 per 100 CAVs); this is where the static obstacles of the roadside LiDAR act (\cref{sec:exp:abl}).
\begin{table*}[t]
\centering\small
\caption{MDrive, infractions per 100 CAVs, a CAV counted once per kind. Lay: collisions with static objects; Off: off-road; Blk: blocked or timed out. V2XPnP: 67 Real2Sim scenarios from real V2X logs, $\approx$2 CAVs each.}
\label{tab:mdrive}
\begin{tabular}{llcccccc}
\toprule
Subset & Method & \ds$\uparrow$ & Ped & Veh & Lay & Off & Blk \\
\midrule
\multirow{2}{*}{Interaction} & Planner alone & 76.9 & 0.0 & 5.7 & 5.7 & 6.3 & 33.8 \\
 & \ours{} & \textbf{80.5} & 0.0 & 7.0 & 0.9 & 11.5 & 34.0 \\
\midrule
\multirow{2}{*}{InterDrive} & Planner alone & 86.0 & 0.0 & 7.1 & 3.8 & 3.2 & 20.5 \\
 & \ours{} & \textbf{90.3} & 0.0 & 5.1 & 0.0 & 3.8 & 17.3 \\
\midrule
V2XPnP & \ours{} & 87.3 & 1.7 & 6.7 & 0.8 & 4.2 & 5.9 \\
\bottomrule
\end{tabular}
\end{table*}

\section{Additional results}
\label{sec:suppl_more}
\begin{figure*}[t]
\centering
\includegraphics[width=\textwidth]{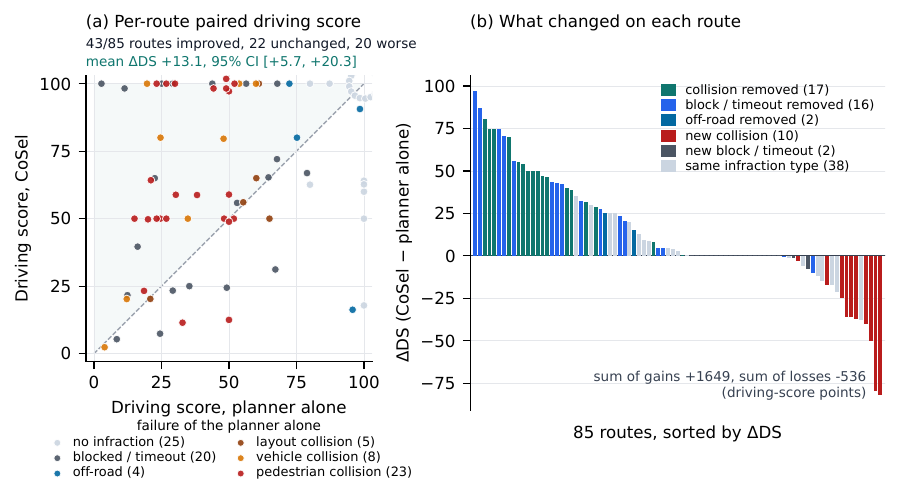}
\caption{Per-route paired driving score on V2Xverse-85. (a) Planner alone vs.\ planner with \ours{}; point color is the planner's dominant failure. Points at (100,100) are spread in a small ring for visibility. (b) $\Delta$\ds{} per route, colored by what changed.}
\label{fig:paired85}
\end{figure*}

\begin{figure*}[t]
\centering
\includegraphics[width=\textwidth]{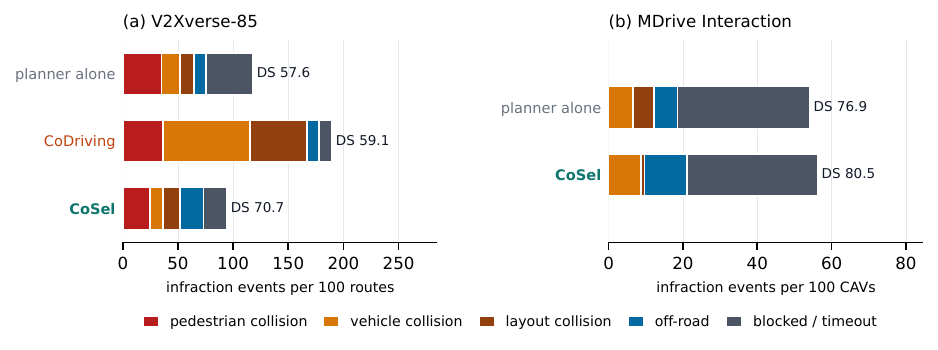}
\caption{Infraction events per 100 routes (V2Xverse-85) and per 100 CAVs (MDrive Interaction), with \ds{} at the bar end. Compared with the planner alone, CoDriving has fewer blocked routes and more vehicle and layout collisions; \ours{} has fewer collisions and fewer blocked routes on V2Xverse-85.}
\label{fig:failbars}
\end{figure*}

\begin{figure*}[t]
\centering
\includegraphics[width=\textwidth]{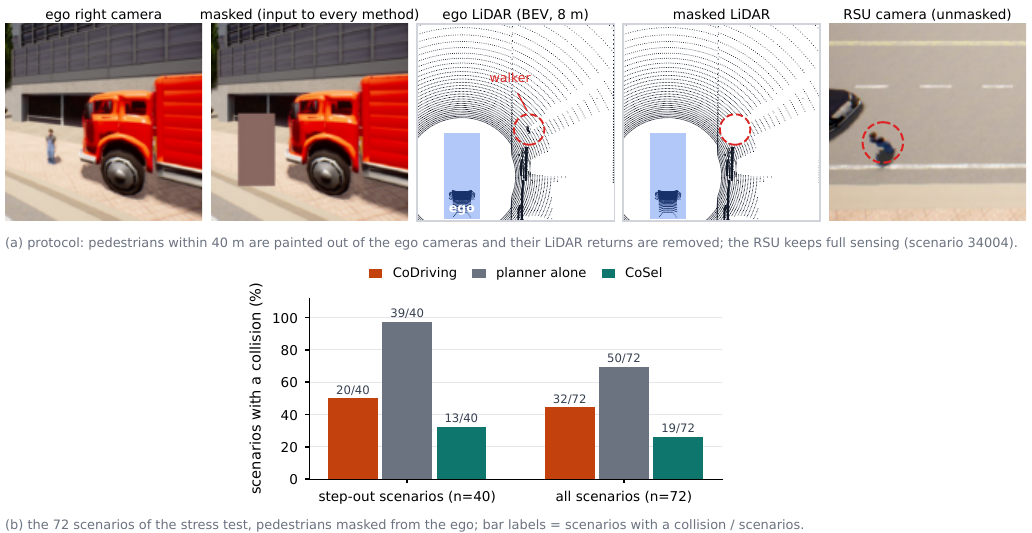}
\caption{Pedestrian sensor mask. (a) The pedestrians are painted out of the ego cameras and their returns are removed from the ego LiDAR; the RSU is unchanged. (b) Scenarios with a collision for CoDriving, the planner alone and \ours{} with the pedestrians masked.}
\label{fig:mask_full}
\end{figure*}

\begin{figure*}[t]
\centering
\includegraphics[width=\textwidth]{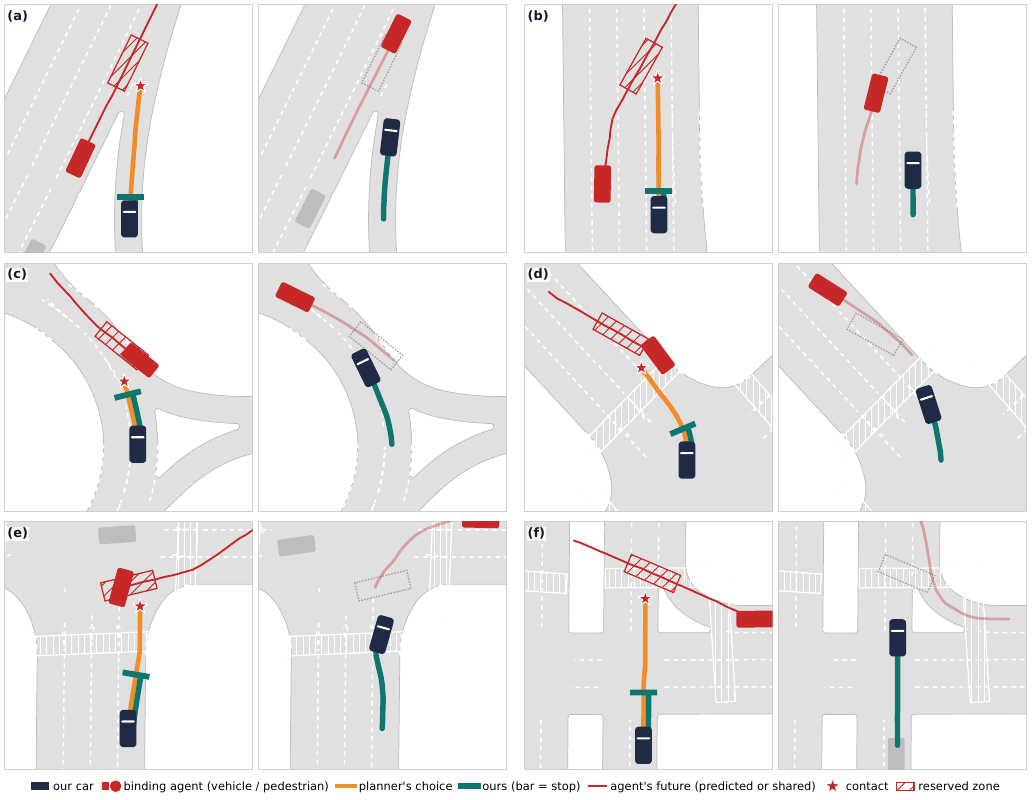}
\caption{Right of way between CAVs on MDrive, six pairwise conflicts. Left: the planner's choice (orange) of the waiting CAV (navy) would enter the conflict zone reserved for the CAV with priority (red), and the selected candidate (green) stops before it. Right: the waiting CAV goes after the other has passed.}
\label{fig:multicav_full}
\end{figure*}

\begin{figure*}[t]
\centering
\includegraphics[width=\textwidth]{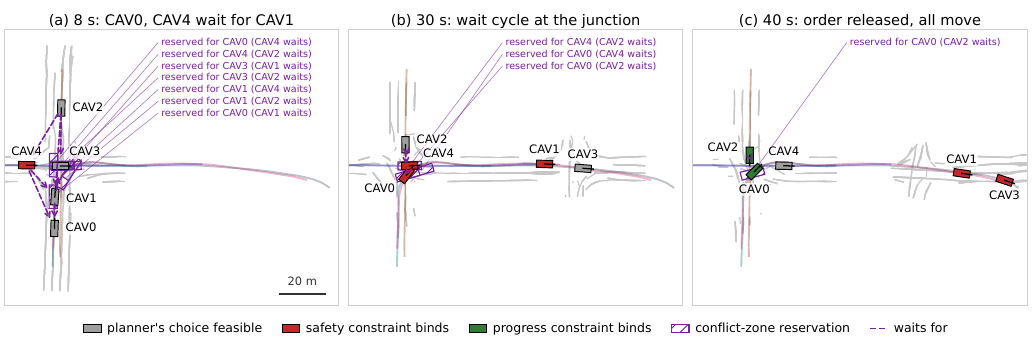}
\caption{MDrive Interaction, Unprotected Left Turn/2, five CAVs: the order over time. In (b) the CAV with priority is itself held by a CAV that yields to it (labeled wait cycle: the order has no cycle, the two hold each other through the order and through the road). The yielding CAV requests a change, all CAVs apply it at the next step, and the order stays total (\cref{sec:suppl_method}).}
\label{fig:multicav5}
\end{figure*}

\begin{figure*}[t]
\centering
\includegraphics[width=0.75\textwidth]{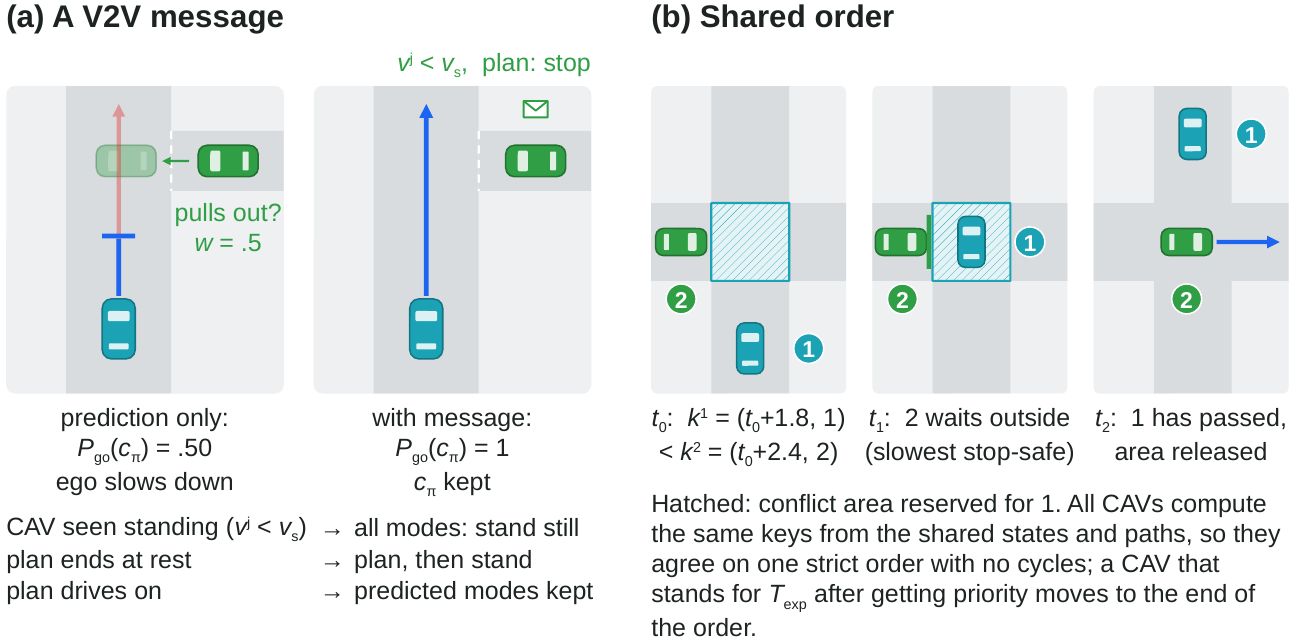}
\caption{(a) A V2V message: a CAV that reports standing, or a plan that ends at rest, replaces the predicted modes of that CAV until its next message, so the ego keeps the planner's choice instead of slowing down. (b) Shared order: every CAV computes the same keys (estimated arrival at the conflict point, CAV index) from the shared states; the lower-priority CAV waits outside the zone reserved for the higher-priority one until it has passed.}
\label{fig:v2v_order}
\end{figure*}

\Cref{fig:paired85,fig:failbars} give the driving score per route and the infractions by kind, \cref{fig:mask_full} the sensor mask of the
stress test, and \cref{fig:multicav_full,fig:multicav5,fig:v2v_order} the right of way between CAVs.

\section{Additional analysis of the cooperative channel}
\label{sec:suppl_channel}

\paragraph{Stress test.}
The \emph{stress test} is a hard diagnostic set that we built so that cooperative information is decisive. Each of its 72 short scenarios is staged next to a fixed RSU rig (four $100^\circ$ cameras, pitch $-40^\circ$, 7.5\,m high) that covers the conflict area: in 40 scenarios (R1) a pedestrian (adult, child, elderly or jogger) steps out from behind a kiosk, a parked truck or a van, 1.4--5.0\,s before reaching the ego's path, in clear, rainy or sunset light; in 5 (R2) a vehicle enters a junction from behind an occluder; 3 lane-change conflicts (R4), and 24 controls (R5): 8 with a crossing that the ego sees, 8 with a pedestrian who walks
away along the sidewalk and never enters the road, and 8 without a pedestrian. When a pedestrian starts to walk toward the ego's path, the ego is a median of 12\,m from the crossing point in the stress test, against 35\,m on V2Xverse-85 and 20\,m on MDrive. In the setting used for the analysis in \cref{sec:exp:channel}, the pedestrians are additionally removed from the ego's cameras and LiDAR, so that the RSU is the only evidence about them: without the RSU boxes, the planner alone and our selection on ego evidence collide in 39 and 38 of the 40 dart-outs. The set is used only to diagnose the cooperative channel; no parameter is chosen on it.
\emph{Remaining collisions.} With full ego sensing \ours{} collides in 17 scenarios, 13 with a pedestrian and 4 with a vehicle.
The 40 step-out scenarios differ in the time between the moment the pedestrian becomes visible to the ego and its entry into the ego's lane:
in 18 it is 1.0--1.5\,s (median 1.0\,s), in 22 it is 1.8--2.7\,s. A full stop from the approach speed of 5\,m/s takes 1.1\,s with a reaction time
of 0.3\,s and a deceleration of 6\,m/s$^2$. Nine of the 11 step-out collisions of \ours{} fall in the first group and 2 in the second.
In 14 of the 17 collisions the planner's choice had exceeded the risk budget within the 3\,s before the contact, so the selector had acted.
The pedestrians follow a scripted motion and do not react to the ego: in 9 of the 13 pedestrian collisions the pedestrian meets the right
side of the moving ego, and in 4 the front of the ego after it has braked to below 0.5\,m/s.

\begin{figure}[t]
\centering
\includegraphics[width=0.72\linewidth]{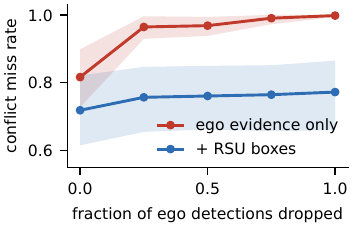}
\caption{Open-loop replay on the stress test: conflict miss rate against the fraction of ego detections dropped, with ego
evidence only and with the RSU boxes added. Bands: bootstrap 95\% intervals over scenarios.}
\label{fig:robust}
\end{figure}

\paragraph{Open-loop: ego-detection loss.}
On the recorded scenarios of the stress test we replay the selection with each ground-truth track removed from the ego detections with probability $p$ (three seeds, bootstrap by scenario), with and without the RSU (Fig.~\ref{fig:robust}). We report whether a tick on which the planner's own trajectory is in conflict is still accepted (risk miss), whether the selection is conflict-free on those ticks (safe pick), and whether a free tick is stopped (over-stop). With the ego fully blind ($p=1$) the risk miss is 0.77 $[0.66,0.87]$ with the RSU and 1.00 without; the safe pick is 0.22 with the RSU and 0.00 without (0.29 vs.\ 0.21 at $p=0$); over-stop stays at 0.19--0.20 in every condition, so the RSU does not buy safety with extra stopping. Tick-level misses are high in every condition because a conflict 2\,s ahead can still be avoided by braking later, so the selection keeps the planner's choice until its stop branch would no longer avoid it; the scenario-level view is the fairer one: on the 30 R1 scenarios with a ground-truth conflict, the first flag precedes the expected overlap by a median of 1.6\,s with the RSU (3 of 30 unflagged) against 1.2\,s with ego detections only (5 unflagged); a blind ego with the RSU keeps 1.4\,s, a blind ego without it flags 3 of 30.

\section{Method details}
\label{sec:suppl_method}
This section gives the definitions that \cref{sec:method} refers to.

\paragraph{Positives of the planner loss.}
The tests of \cref{sec:planner} run over the planner's horizon of 2\,s, with $\eta_s=1$\,m, $\eta_v=0.5$\,m/s and $\eta_\tau=0.5$\,m.
Mean speed differences and mean position errors are taken over the valid steps of the expert's future, and a collision is tested
against the boxes of the logged agents.
If no profile within $\eta_s$ passes the speed test, the one with the smallest speed difference is the positive, and if no composed
candidate passes, the nearest one without a collision is.
Below an ego speed of 0.3\,m/s the traveled distance does not separate starting now from starting later, and only the positives that
agree with the expert's action, to brake or to drive, are kept: when the expert brakes, the profiles whose mean speed over 0.4--0.8\,s
is at least 0.3\,m/s are removed, and when it drives, only these profiles are kept.
The positives of a frame share the weight of one positive.

\paragraph{Predictor training.}
Each detected box of the present and the three previous decision steps (0.6\,s) is a token with its position, heading, size, class, score,
source, age and velocity, in the present ego frame; a plan of a CAV over 2\,s is one more token.
All CAVs decide at the same step, so the newest plan of a CAV is the candidate that it chose at the previous step and is executing;
the ego places this candidate at the CAV's present pose.
Only a plan that ends in a stop is taken as given: the CAV can always brake, so it goes no farther than the end of such a plan, and if
other agents hold it up, it stops earlier on the same path, which its next message shows.
We take such a message as given instead of predicting the CAV for three reasons.
The predictor gives each agent a marginal future and would spread probability over futures that the CAV has already ruled out by its
own decision; a stationary CAV or a stop plan fixes the CAV over the whole horizon $H$ by the CAV's own action alone, so the ego can
plan around it as around a static obstacle; and the message is replaced at the next decision step, so the assumption lasts one step.
The interaction between CAVs, which this does not model, enters through their shared order and the conflict zones of
\cref{sec:selection}; a plan that ends in motion depends on that interaction and stays an input of the predictor.
Three self-attention layers of width 128 (0.53\,M parameters) output, for every present box, $M=6$ trajectories over $H=4$\,s in steps of
0.2\,s as offsets from constant velocity, their probabilities and the existence $e_i$.
A box within 2\,m of a logged agent has existence target 1 and that agent's future as target; other boxes have existence target 0.
The loss is smooth-$\ell_1$ on the mode nearest to the target, with cross-entropy on that mode and binary cross-entropy on $e_i$, both
weighted 0.5.
Training removes the ego's boxes of an agent that another source also sees with probability 0.5, all RSU boxes with 0.2 and all CAV boxes
and plans with 0.3; a plan is given with probability 0.5.
We train four epochs on 63k frames from Towns 1--4 and 6 with AdamW (learning rate $5\times10^{-4}$, batch size 32); Towns 7 and 10 are
used for validation.

\paragraph{Boxes of the same agent.}
The predictor receives every box.
Afterward a box from an RSU or a CAV within 1.5\,m (pedestrians 0.8\,m) of an ego box of the same class is removed, and CAVs send no
detections of other CAVs.
Two remote boxes of an agent that the ego does not detect are not merged: each is a factor of \cref{eq:pb}, so the predicted risk of that agent
is counted twice. This can occur on MDrive and not on V2Xverse-85, which has one remote source.

\paragraph{Walking pedestrians.}
A pedestrian counts as walking when its own detections one or two decision steps apart give a speed of at least 0.5\,m/s, unless it is
behind the ego and faces the ego's direction of travel.
It is represented by a disc around the center of its box that keeps the observed velocity over $H$; the radius is the larger half extent of the
box plus the distance the pedestrian covers during the committed prefix $T_p$, because it may be one decision ahead of the extrapolated position.
A candidate must stay clear of this disc when it goes on; a later stop of the ego gives no credit against it.
The predicted modes of a walking pedestrian stay in \cref{eq:pb} as well.
Cyclists, vehicles and pedestrians that do not walk enter with their predicted modes only; an agent whose modes give an expected speed below 0.5\,m/s is checked
as one standing box, with a margin for the tracking error of the ego.

\paragraph{Start corridor.}
The corridor applies where no roadside LiDAR is available, for an ego at rest whose planner chooses to move or for which the progress
rule applies, and while the ego is less than $D_s=10$\,m from where it last stood, the spacing of the navigation points.
Its edges lie one lane width (3.5\,m) plus the footprint clearance to either side of the straight line from the ego to its current
navigation point.
The corridor uses a single navigation point, the one the planner itself receives.

\paragraph{Probability of a safe continuation.}
Every candidate has two continuations: to go on with its speed profile, and to follow it for the committed prefix $T_p=0.4$\,s (one
decision period plus one actuation step) and then brake to a standstill at $a_b=3$\,m/s$^2$.
Let $s_c^{\text{go}}(i,m)$ and $s_c^{\text{stop}}(i,m)$ indicate that the continuation has no contact with mode $m$ of box $i$ within
$H$.
In the stop continuation a contact while the ego still moves always counts, and a contact with the standing ego counts unless the
agent stands or is obliged to stop for it and can do so (reaction time 0.5\,s, deceleration 4\,m/s$^2$ for vehicles and 2\,m/s$^2$ for
pedestrians); obliged are agents that start behind the ego and move in its direction and CAVs that yield to the ego in the order.
With $a_i^{x}=(1-e_i)+e_i\sum_m w_{im}\,s_c^{x}(i,m)$ for $x\in\{\text{go},\text{stop}\}$ and $a_i^{\text{both}}$ the same sum over
$s_c^{\text{go}}s_c^{\text{stop}}$,
\begin{equation}
  P_c^{\text{adm}}=\prod_i a_i^{\text{go}}+\prod_i a_i^{\text{stop}}-\prod_i a_i^{\text{both}},
\end{equation}
the probability that at least one continuation stays without contact, with $P_c^{\text{go}}=\prod_i a_i^{\text{go}}$ as in \cref{eq:pb}.
$P_c^{\text{adm}}$ counts a future as safe when either continuation is safe in it, so it is the probability that a safe continuation exists and an
upper bound on the probability that a continuation fixed in advance is safe; the two differ when the continuations are safe in different futures.
The selector decides again every 0.2\,s with the new detections, and the stop continuation of a kept candidate begins only after the committed prefix.
The stricter rule, which keeps a candidate only if going on is safe, is the last row of \cref{tab:ablation}.

\paragraph{Progress.}
The stall case applies when the planner's choice has stood for $T_{\text{stall}}=4$\,s, the ego has moved less than 2\,m in that window
and, at every step of the window, a moving candidate on the planner's path was in $\mathcal{G}$.
The selector then executes the moving candidate on that path with the highest planner score, until the planner's own choice moves.
The blocked-path case applies when a static obstacle lies on $p_\pi$ within $D=40$\,m.
The vocabulary paths whose footprint stays clear of static obstacles join $\mathcal{C}$, and among those with a moving candidate in
$\mathcal{G}$ the selector takes the path that advances furthest along the navigation points; without one, the ego stops before the
obstacle.

\paragraph{Constants.}
The committed prefix $T_p=0.4$\,s is one decision period plus one actuation step, and the stall window $T_{\text{stall}}$ and the time-out
$T_{\text{exp}}$ are 4\,s, as long as the horizon.
The start distance $D_s=10$\,m is the spacing of the navigation points, and the look-ahead distance $D=40$\,m is more than the ego
covers within $H$ (32\,m at the speed limit of 8\,m/s on V2Xverse).
The risk budget was chosen on validation scenarios (\cref{sec:suppl_delta}).

\paragraph{Order between CAVs.}
Two CAVs are in conflict when their paths ahead, up to $D$, come within $r_c=2.5$\,m of each other and neither follows the other.
The key of a CAV is its estimated arrival time at the conflict, then its index, so the order is total and has no cycle, and every pair
holds mirrored relations: one has priority, the other yields.
A CAV with priority can still be unable to pass, and its priority over a yielding CAV ends in the following cases, each read from the shared
states: it has stood for $T_{\text{exp}}=4$\,s since it received the priority while the yielding CAV also stands; the yielding CAV is already
within $r_c$ of the path ahead of the CAV with priority, where waiting blocks that CAV; neither has made progress for $T_{\text{exp}}$ and
the CAV with priority has no candidate in $\mathcal{G}$ or names the yielding CAV as the agent that makes its choice unsafe; or the CAV with
priority has stood for $T_{\text{exp}}$ with a plan that ends before the conflict, or has not moved for $T_{\text{exp}}$ although the candidates it selected move.
The yielding CAV then sends a request, and at the next decision step every CAV applies the same change: the CAV that had priority is
placed directly behind the requester, or at the end of the order when it has stood for $T_{\text{exp}}$.
The order stays total.
Requests of, and against, a CAV that has reached its destination are discarded.
A yielding CAV that stands, with $\mathcal{G}$ empty, takes the slowest candidate in $\mathcal{A}$ and does not creep toward the
conflict zone.

\section{One decision step}
\label{sec:suppl_alg}
At every decision step (0.2\,s) the selector does the following.
\begin{enumerate}
  \item \emph{Candidates.} $\mathcal{C}$ is the planner's path $p_\pi$ and its $K$ highest-scored paths, each with the speed profiles
  that the planner keeps after its first scoring round, plus the stop profile. If a static obstacle blocks $p_\pi$ within $D$, the
  vocabulary paths that pass it are added. The planner's network, weights and vocabulary are not changed.
  \item \emph{Scene.} The predictor gives existence probabilities and future modes for the detections of the ego, the RSUs and the
  CAVs. The given items are added with $e_i=1$ and one mode: CAVs that stand or whose shared plan ends in a stop, walking
  pedestrians, static cells, solid lane lines, the start corridor, and the conflict zones of CAVs with priority.
  \item \emph{Probabilities.} For every candidate, $P_c^{\text{go}}$ (\cref{eq:pb}) and $P_c^{\text{adm}}$, the probability that going
  on or braking after the committed prefix stays without contact; $\mathcal{G}$ and $\mathcal{A}$ are the candidates at or above
  $1-\delta$.
  \item \emph{Safety.} The key of \cref{eq:select}: a candidate in $\mathcal{A}$ before one outside it, one in $\mathcal{G}$ before one
  outside it, then the smallest departure from the planner's choice. A CAV that yields and stands, with $\mathcal{G}$ empty, takes the
  slowest candidate in $\mathcal{A}$. If $\mathcal{A}$ is empty, the candidate with the lowest speed after the committed prefix.
  \item \emph{Progress.} If the planner has chosen to stand for $T_{\text{stall}}$ and the ego has moved less than 2\,m in that window, while a
  moving candidate on its path stayed in $\mathcal{G}$, that candidate with the highest planner score. If a static obstacle blocks $p_\pi$ and a moving candidate on a
  passing path is in $\mathcal{G}$, the passing path that advances furthest along the navigation points.
  \item \emph{Execution.} The selected candidate is executed. At every step the log holds the agents with the smallest factors of
  $P_{c_\pi}^{\text{go}}$ and the source of each; an agent that the ego and an RSU both detect counts as the ego's. For the given
  items the log holds how many candidates each kind removed.
\end{enumerate}
Two statements of the main text are separate. The planner is never changed, in any case. The executed candidate is the planner's
choice whenever that choice is in $\mathcal{G}$, with the two progress cases of step 5 as the only exceptions.

\section{Failure causes on V2Xverse-85}
\label{sec:suppl_fault}
\begin{figure*}[t]
\centering
\includegraphics[width=\textwidth]{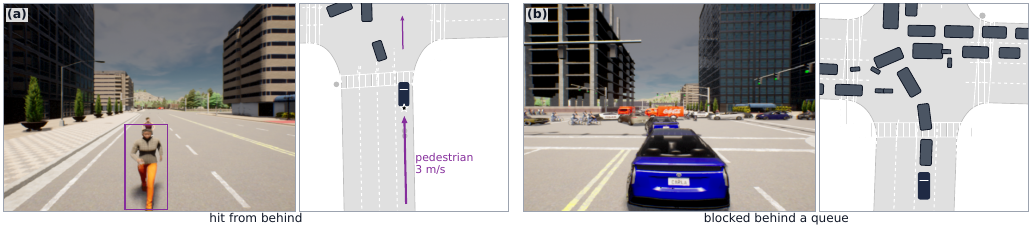}
\caption{\textbf{Two failures of \ours{} on V2Xverse-85}, each as a camera frame and a bird's-eye view of the same instant.
(a) Rear camera, 1.3\,s before a contact: the pedestrian that the scenario places behind the ego runs along the lane at 3\,m/s and reaches the ego,
which moves at about 0.5\,m/s (star: contact). The planner alone is also hit from behind by a pedestrian on this route.
(b) Front camera at the end of a blocked route: the ego stands behind two standing cars in its lane, and the junction ahead is filled with standing
vehicles (dark gray). The planner alone ends blocked at the same place.}
\label{fig:fault_cases}
\end{figure*}

\Cref{tab:fault} splits the collisions with agents on V2Xverse-85 by where the struck agent was at the impact, in the frame of the ego,
and by the speed of the ego, and the blocked routes by what stood in front of the ego during its last ten seconds.
We can make this split for our own runs, which log the ground-truth agents; the logs of the baselines do not contain them.
One scenario type of the benchmark places a pedestrian in the lane behind the ego who runs along the lane at 3\,m/s and reaches any
car that slows down.
\Cref{fig:fault_cases} shows one collision from behind and one blocked route.
\begin{table}[h]
\centering\footnotesize
\caption{\textbf{V2Xverse-85, collisions with agents and blocked routes.} Behind: the struck agent is in the rear sector of the
ego at the impact. Standing: not from behind, ego below 0.5\,m/s. Moving: not from behind, ego at or above 0.5\,m/s. Agent ahead: a
standing or slow agent is in front of the ego during the last 10\,s of a blocked route. In a queue: two or more standing agents in the
ego's lane within 40\,m.}
\label{tab:fault}
\setlength{\tabcolsep}{2.2pt}
\begin{tabular}{l ccc ccc}
\toprule
 & \multicolumn{3}{c}{Planner alone} & \multicolumn{3}{c}{\ours{}} \\
\cmidrule(lr){2-4}\cmidrule(l){5-7}
Collisions & behind & standing & moving & behind & standing & moving \\
\midrule
All agents & 17 & 8 & 19 & 10 & 9 & 12 \\
\midrule
Blocked routes & \multicolumn{3}{c}{36} & \multicolumn{3}{c}{17} \\
\quad agent ahead & \multicolumn{3}{c}{25} & \multicolumn{3}{c}{13} \\
\quad \quad of these in a queue & \multicolumn{3}{c}{15} & \multicolumn{3}{c}{7} \\
\quad nothing ahead & \multicolumn{3}{c}{11} & \multicolumn{3}{c}{4} \\
\bottomrule
\end{tabular}
\end{table}
Against the baselines, all on the same 85 pairs, the paired \ds{} difference of \ours{} is $+11.6$ $[+3.7,+19.6]$ to CoDriving, $+7.5$ $[+0.3,+14.7]$ to late fusion,
$+19.9$ $[+13.0,+27.1]$ to F-Cooper and $+18.7$ $[+10.6,+27.0]$ to V2X-ViT, and a collision occurs on 32 routes against 52, 52, 64 and 59.
TCP completes 44 of the 85 routes and \ours{} 67.
Against the planner alone, \ours{} has 7 fewer collisions in which the moving ego drives into an agent (19 and 12) and 7 fewer agents that reach the
ego from behind (17 and 10).

\section{Calibration of the predictor}
\label{sec:suppl_calib}
We evaluate the predictor of \ours{} on the validation Towns 07 and 10 (22 routes, 5{,}962 frames), with the fusion of the deployed
system, in three evidence settings: ego detections only, ego and RSU detections, and all sources with shared plans.
The expected calibration error of the existence probabilities is 0.038, 0.026 and 0.031, and that of the mode weights 0.013, 0.008
and 0.009.
For the risk of a trajectory we roll out 15 constant-speed probe trajectories per frame and compare $1-P_c^{\text{go}}$ with the
frequency of a contact with the logged agents: the error is 0.074, 0.073 and 0.079.
The risk is overestimated in the middle range. With ego and RSU detections, a predicted risk of 0.35 corresponds to an observed
frequency of 0.24, 0.65 to 0.46 and 0.75 to 0.57, so the budget $\delta=0.3$ corresponds to an observed risk of about 0.2.
At the low end, shared evidence removes missed risk: of the trajectories with a predicted risk of zero, 7.1\% hit an agent with ego
detections only, 2.9\% with the RSU detections and 2.4\% with all sources.

\section{Earlier comparisons with a previous planner}
\label{sec:suppl_alt}
Before \ours{} we tried three other ways of using the shared information.
The first two used an earlier version of our planner (same architecture, other weights), and the third partly an earlier checkpoint of
the predictor, so their numbers are not comparable with the tables of the main text; we report them because they motivated the design.

\paragraph{RSU features fused into the planner.}
We fused the RSU detections into the planner in two ways: through a cooperative perception head whose fused queries the planner
reads, and as additional tokens of the planner's agent attention.
\emph{Frozen planner.} With the fused queries, the average displacement error on 10{,}208 held-out frames falls by 0.026\,m
$[0.020, 0.032]$, but the planner slows less often when the expert brakes ($-4.8$ points $[-6.8,-3.0]$).
With the additional tokens, the speed of the planner's choice changes as much with the RSU tokens of another frame as with the real
ones (paired difference $-0.005$\,m/s).
\emph{Adapter.} A 131\,k-parameter adapter on the RSU tokens, trained on the full training set with the planner frozen, slows the
planner wherever tokens are present; trained on real tokens or on shuffled tokens it behaves the same on the frames in which the RSU
sees the cause of the expert's braking (paired difference of the chosen speed $-0.028$\,m/s $[-0.102,+0.037]$).
\emph{Joint training.} A planner fine-tuned on ten routes with the fused queries is not better than one fine-tuned without them and
given them only at test time (displacement error 0.696\,m against 0.682\,m), and on the 93 frames in which only the RSU sees the
cause of the expert's braking it slows in 38\% with the RSU and 42\% without.
\emph{Closed loop.} On V2Xverse-85 the planner trained with fused RSU features does not improve on the same planner without them
(paired \ds{} difference $-6.7$ $[-13.0,-0.1]$); the two also differ in their training schedule.
\emph{Why.} The expert data of this benchmark explain this: with the RSU 12\,m ahead of the ego, among the 7{,}158 braking events of the expert there is no sustained one whose cause is an
agent ahead that only the RSU sees (\cref{sec:suppl_decisive}), so no training sample requires the fused features, and shuffling them at test time changes the planned speed by 0.005\,m/s.
Fusion can learn such a response from data constructed to contain it~\cite{coopernaut}; selection does not depend on such data.

\paragraph{A learned re-ranker of the candidates.}
We trained a scorer that reads the candidates of the frozen planner and the detected boxes and predicts a contact within 2\,s, on all
119{,}857 frames of the training set.
Its AUROC on held-out frames is 0.945, and 0.958 with the velocity of the tracked boxes as input.
On 19 closed-loop collisions with a pedestrian that steps into the lane, it flags the planner's choice one to two seconds before the
impact in 1 case from the boxes alone and in 9 with RSU boxes and their velocity.
Re-ranking the final candidates avoids 1 of the 19: in 18, the planner had removed every safe speed profile in its first scoring
round.
A hard veto in that first round removed candidates at 49\% of the decision steps in closed loop, and the route completion fell from
93.6 to 74.5 on the first six routes, where we stopped the run.
The comparison shows that the candidates that reach the selection matter: the scorer worked on the final candidates, from which the safe
speed profiles were already gone. The candidates of \ours{} therefore come from before the planner's final round and always include the stop
profile. The comparison does not show whether predicting the motion of the agents is better than predicting the contact of a candidate.

\paragraph{Constant-velocity futures.}
\emph{Validation towns.} \Cref{tab:suppl_pred} compares the predictor with a constant-velocity extrapolation of the same boxes on 5{,}962 frames of the validation Towns 07 and 10.
The risk is evaluated on 89{,}430 probe trajectories of 4\,s and on the planner's candidates in 945 frames with at least one colliding candidate.
Constant velocity has almost three times the forecast error, ranks risky trajectories worse (AUROC 0.83 against 0.93, paired difference $+0.10$ $[0.10,0.11]$) and calls safe three times as many trajectories that hit an agent (8.9\% against 2.8\%).
The most probable mode of the predictor alone has an error close to that of constant velocity (1.31\,m), so the gain comes from the set of modes.
With the ego's tokens only, 7.1\% of the zero-risk probe trajectories hit an agent; the RSU and CAV tokens bring this to 2.8\%.
\begin{table}[h]
\centering\small
\caption{Predictor against constant velocity on the validation towns. Probe: 4\,s trajectories at constant speed; 0-risk hit: share of the trajectories with zero predicted risk that hit an agent.}
\label{tab:suppl_pred}
\setlength{\tabcolsep}{3pt}
\resizebox{\linewidth}{!}{%
\begin{tabular}{l cc cc c}
\toprule
 & minADE & minFDE & \multicolumn{2}{c}{Probe} & Cand. \\
Futures & 2\,s (m) & 4\,s (m) & AUROC & 0-risk hit & AUROC \\
\midrule
Standing boxes & 2.22 & 8.21 & 0.778 & 13.2\% & 0.824 \\
Constant velocity & 1.39 & 5.83 & 0.825 & 8.9\% & 0.774 \\
Predictor & \textbf{0.52} & \textbf{2.02} & \textbf{0.928} & \textbf{2.8\%} & \textbf{0.884} \\
\midrule
True futures & -- & -- & 0.946 & 1.8\% & 0.912 \\
\bottomrule
\end{tabular}}
\end{table}
\emph{Closed loop.} On 53 validation scenarios in Towns 07 and 10 (33 with a crossing pedestrian or cyclist, two runs each) and with the
planner of \ours{}, a hand-written braking rule on the positions and constant-velocity motion of pedestrians and cyclists lowers the
scenarios in which the ego is at fault from 8 and 7 for the planner alone to 4 and 3 with ego detections and to 3 and 0 with the RSU
detections added.
The rule slows the ego on the control scenarios as well (mean route time 33.3\,s against 30.7\,s).

\section{Choice of the risk budget}
\label{sec:suppl_delta}
The risk budget 0.3 was chosen on 53 validation scenarios in Town10 (32) and Town07 (21), none of them in Town05.
\Cref{tab:suppl_delta} repeats the scan with the selector of this paper: the planner-alone run of these scenarios is replayed offline, and at each of
its 8{,}338 decision steps the selector chooses from the logged state.
A step is unsafe when the executed candidate, followed over $H$, touches the logged future of an agent; with the planner's own choice 8.70\% of the
steps are unsafe.
The table also gives how many of the 11 scenarios in which the planner alone has a contact are flagged before the contact, how early, and how often the six
scenarios without a conflict are flagged.
The budget trades the unsafe steps against the flags on scenarios without a conflict, without a knee.
\emph{Other constants.} On the same scenarios we varied nine constants of the selector one at a time: the deceleration of the stop continuation
(2 and 4\,m/s$^2$), the reaction time (0.3 and 1\,s) and the deceleration of the other agents, the radius of the ego footprint ($\pm0.1$\,m), the margin on
standing agents, the number of candidate paths (8 and 32), the acceleration bound of the committed prefix, the standing threshold (0.3 and 0.8\,m/s) and
$T_{\text{stall}}$ (2 and 8\,s).
For the first seven the executed candidate changes at no more than 1.2\% of the steps and the unsafe steps stay between 7.77\% and 7.81\%.
The standing threshold and $T_{\text{stall}}$ change how often the selector moves a standing ego on, and the unsafe steps stay between 7.65\% and 7.78\%.
The replay is open loop; the start corridor, the lane lines, the static obstacles and the order between CAVs have no input on these scenarios.
On V2Xverse-85 the planner's choice has a predicted risk of exactly zero at 63\% of the decision steps and above 0.3 at 21\%.
\emph{Collisions within the budget.} A budget of 0.3 accepts candidates with a predicted risk up to 0.3, and we count the collisions that it let pass.
For every collision of \ours{} with an agent we take the decision steps of the 3\,s before the impact.
On V2Xverse-85 the planner's choice exceeds the budget at one of these steps in 27 of the 31 collisions, so the selector had already acted; in 4 the predicted risk
stays within the budget (at most 0.24), 3 of them with the ego driving into a pedestrian; in none is it zero.
On the stress test with the pedestrians masked the counts are 11, 3 and 5 of 19.
The risk is that of the whole scene, so 4 and 3 are upper bounds of the collisions that a smaller budget could have changed.
\begin{table}[h]
\centering\small
\caption{Validation scenarios (Towns 7 and 10), offline replay of the selector: risk budget $\delta$.}
\label{tab:suppl_delta}
\begin{tabular}{c cccc}
\toprule
$\delta$ & Unsafe & Flagged before & Median & Flagged steps, no- \\
 & steps (\%) & contact (of 11) & lead (s) & conflict scen.\ (\%) \\
\midrule
0.1 & 7.25 & 11 & 5.85 & 11.4 \\
0.2 & 7.58 & 10 & 4.75 & 6.5 \\
0.3 & 7.77 & 9 & 2.30 & 5.4 \\
0.4 & 7.88 & 8 & 1.90 & 3.8 \\
0.5 & 7.95 & 8 & 1.90 & 3.3 \\
\bottomrule
\end{tabular}
\end{table}

\section{The selector on the CoDriving planner}
\label{sec:suppl_host}
To test the selector on a planner that it was not built for, we attach it to the released CoDriving model.
CoDriving plans one path; the candidates are this path with the speed profiles of our vocabulary, and the profile closest to
CoDriving's own waypoints is its choice.
The detections are those of CoDriving, and the predictor, the scene and the selector are unchanged.
With the selector the \ds{} is 65.6 on all 85 routes, against 59.1 for CoDriving with its own controller (\cref{tab:main}), with 39 vehicle
collisions against 67 and 8 blocked routes against 10.
The run with the selector uses our controller, with which CoDriving alone scores 39.4 on the 77 routes it finished (paired difference
$+26.2$ $[+18.7,+33.8]$).

\section{The selector on SparseDriveV2 in Bench2Drive}
\label{sec:suppl_b2d}
We also attach the selector to the released SparseDriveV2~\cite{sparsedrivev2}, trained on the Bench2Drive data of the Think2Drive
expert~\cite{bench2drive,think2drive}, on 45 Bench2Drive routes: 20 with a crossing pedestrian (four scenario types, five routes each) and 25 at
junctions (five types, five routes each).
The candidates are SparseDriveV2's own paths and speed profiles and the detections are its own; the predictor is that of \ours{}, and the
selector is an earlier version without the walking-pedestrian disc, the start corridor and the path case of \cref{sec:selection}.
The candidate speeds are capped at SparseDriveV2's own choice, so the selector can only slow down or change the path, and the speed case
cannot act either: on no route does the selector start a standing ego.
With the ego's detections only, the selector raises the driving score by $+9.2$ $[+2.0,+17.0]$, paired over the 45 routes with two runs of
each route for both systems.

\section{What enters the scene}
\label{sec:suppl_items}
The selector applies one criterion, \cref{eq:problem}, to every candidate.
What differs between the cases of \cref{sec:predictor} is what stands in the scene and with which future (\cref{tab:suppl_items}).
Traffic lights are not in the scene: the planner sees them through its cameras, and the stop line of a red light can be added as one
more given item.
The progress term adds no traffic infraction on V2Xverse-85: no method runs a red light on these routes, and \ours{} commits as many stop-sign
infractions as its planner (8; CoDriving 10).
\begin{table}[h]
\centering\small
\caption{Items of the predicted scene.}
\label{tab:suppl_items}
\setlength{\tabcolsep}{3pt}\begin{tabular}{p{0.30\linewidth} p{0.22\linewidth} p{0.38\linewidth}}
\toprule
Item & Source & Future over $H$ \\
\midrule
Detected agent & ego, RSU, CAV & predictor, $M$ modes with existence $e_i$ \\
CAV that stands or plans a stop & its message & stays, or follows its plan and stays \\
Walking pedestrian & ego, RSU, CAV & keeps its observed velocity \\
Static obstacle & roadside LiDAR & stays \\
Solid lane line & online map & stays \\
Start corridor & navigation point & stays while the ego starts \\
Conflict zone of a CAV with priority & CAV messages & occupied until that CAV has passed \\
\bottomrule
\end{tabular}
\end{table}

\end{document}